\documentclass[10pt,twocolumn,letterpaper]{article}
\usepackage[pagenumbers]{cvpr}

\usepackage{cuted} %

\usepackage{currfile} %

\usepackage{caption} %

\usepackage{booktabs}
\usepackage{colortbl} %
\AtBeginDocument{\definecolor{bestcell}{RGB}{214,230,248}\definecolor{gaincolor}{RGB}{30,130,60}\definecolor{losscolor}{RGB}{200,40,40}}
\usepackage{graphicx}
\usepackage{amssymb}
\usepackage{amsmath}
\usepackage{multirow}
\usepackage{url}

\definecolor{cvprblue}{rgb}{0.21,0.49,0.74}
\usepackage[pagebackref,breaklinks,colorlinks,allcolors=cvprblue]{hyperref}
\def\confName{CVPR}
\def\confYear{2027}

\providecommand{\bench}{Freeze}
\providecommand{\cam}{CamBit}
\providecommand{\scene}{SceneBit}

\title{Camera and Scene Motion Interfere with Each Other \\in Video Representations}
\author{Ziqi Zhou \qquad Laura Sevilla-Lara\\
School of Informatics, University of Edinburgh}
\makeatletter
\patchcmd{\@maketitle}{\vspace*{12pt}}{\vspace*{4pt}}{}{}
\patchcmd{\@maketitle}{\vspace*{24pt}}{\vspace*{14pt}}{}{}
\makeatother
\begin{document}
\maketitle
\providecommand{\bench}{Freeze}
\providecommand{\cam}{CamBit}
\providecommand{\scene}{SceneBit}
\begin{abstract}
Real-world videos often contain both camera motion and scene motion, which are typically encoded jointly within video representations. This joint encoding raises the possibility that the two sources of motion may interfere with one another, potentially affecting the quality of the learned representations. In this paper, we first investigate whether and to what extent camera motion and scene motion adversely affect the performance of state-of-the-art video foundation models on downstream tasks. To this end, we introduce \bench{}, a synthetic benchmark composed of triplets of video clips: one where both objects and camera are moving, the same with fixed camera, and the same with fixed scene. To our knowledge, it is the first benchmark that makes the interference of camera motion and scene motion measurable under controlled conditions. We test four video foundation models on \bench{} and find that all of them perform better on every camera and scene
task once the other motion is frozen, so the motion interference exists in both directions. Given this observation, our second focus is to explore whether the representations of the frozen versions can be approximated simply, in a post hoc manner. We leverage \bench{} to learn two lightweight mappings that take as input video features and bring them closer to the frozen-scene and frozen-camera versions, respectively. The resulting camera representation, \cam{}, improves camera understanding and geometric grounding on real videos for
nearly every task and backbone.
The scene representation, \scene{}, improves the scene tasks of \bench{} and those on real videos, and our analysis
shows that its effect is highly context-dependent.
Overall, \bench{} and the post hoc mappings open a new path towards understanding how
camera and scene motion interact in video representation learning, and mitigating their interference where it arises.
Project page: \url{https://jia1018.github.io/Freeze/}.
\end{abstract}

\providecommand{\bench}{Freeze}
\providecommand{\cam}{CamBit}
\providecommand{\scene}{SceneBit}
\section{Introduction}
\label{sec:intro}

\begin{figure}[t]
  \centering
  \includegraphics[width=\columnwidth,trim=0 0 0 8.6bp,clip]{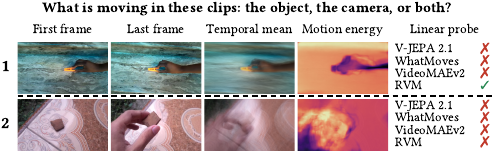}  %
  \vspace{-8pt}
  \caption{Two Something-Something v2~\cite{ssv2} clips, \textit{Pushing something from right to left} (1) and
  \textit{Moving something up} (2), in which both the object and the camera move. Despite the classes being simple, state-of-the-art video backbones misclassify the clips.}
  \label{fig:teaser}
\vspace{-6pt}  %
\end{figure}

Many real-world videos contain a combination of camera motion and scene motion. This includes videos captured by handheld
devices~\cite{ssv2,ling2024dl3dv}, egocentric cameras~\cite{Damen2022RESCALING,grauman2024egoexo4d}, or moving cameras used in
filmmaking~\cite{bain2020condensed,lin2025camerabench,zhou2026unveiling}. In some cases, these two motions can potentially be confused with
each other in a video representation. For instance, consider the two clips in Fig.~\ref{fig:teaser}, in which both the object and the camera move. The frames show only
the combination of the two motions, and the object barely changes its position in them, although it moves in the
scene. Indeed, linear probes on three of four off-the-shelf
backbones~\cite{vjepa21,fundel2026moves,videomaev2,recurrentvideo} misclassify both clips. A possible reason is that
video foundation models learn only from such mixed motion, which may limit how well their representations capture each
of them. In other words, the two motions may interfere with each other in these representations, but whether and how
much they do has not been measured.

The first focus of this paper is to ask the question: \emph{do camera and scene motion interfere with each other in video
representations, and if so, how much?} To address this question empirically, we create a benchmark of triplets of videos
that are nearly identical: a first video with both camera and scene motion, a second video with the same scene motion but
a still camera, and a third one with the same camera motion but a frozen scene. We call this benchmark \emph{\bench}. To
our knowledge, it is the first to compare representations under controlled conditions, keeping everything the same but
switching scene and camera motion on or off.
Built with an existing data generator and dataset~\cite{kubric,huang2025spacetimepilot}, it covers two kinds of scene
motion: object motion driven by physics and humans moving. We evaluate both in camera tasks, such as classifying camera
movements~\cite{lin2025camerabench}, and in scene tasks, such as object physics and the timing of human motion.

Using \bench{}, we test four state-of-the-art video foundation
models~\cite{vjepa21,videomaev2,recurrentvideo,fundel2026moves} and we find the surprising observation that all of them perform significantly better when one of the sources of motion is frozen. This pattern is
consistent across tasks and video representations, and in fact it occurs in both directions.

Motivated by this observation, the second focus of this paper is to ask whether this interference can be mitigated in
a post hoc manner, to avoid the costly re-training of the foundation model. To this end, we learn two lightweight
mappings that take as input the original features of a video clip and map them to a~\emph{\cam} representation, closer
to that of the scene-frozen version of the video, and to a~\emph{\scene} representation, closer to that of the camera-frozen version. These mappings are trained using
the synthetic triplets from \bench{}, as well as real-world videos with simple pretext tasks, such as predicting the camera motion~\cite{zhou2026unveiling} or future prediction.
Experiments show that both mappings improve over the original features on nearly every task and backbone. We also provide an in-depth
analysis of camera freezing approximation, which reveals a more complex, context-dependent problem.

In summary, we make the following contributions:
\begin{itemize}
    \item A benchmark which we call \emph{\bench}, the first to measure empirically how much camera and scene motion
    interfere with each other in video representations.
    \item An analysis showing that this interference exists in both directions, consistently across tasks and video
    foundation models.
    \item Two lightweight mappings that plug into fixed video backbones without fine-tuning them and improve their
    representations of both camera and scene motion.
    \item A first step towards mitigating this interference post hoc, with insights into the scene direction that
    provide a basis for future research.
\end{itemize}

\section{Related Work}
\label{sec:related}

\noindent\textbf{Self-Supervised Video Representations.}
Recent advances in self-supervised video learning have enabled models to capture rich spatiotemporal information without requiring task-specific supervision~\cite{qian2021cvrl,feichtenhofer2021large}. Masked video modeling approaches, such as VideoMAE~\cite{tong2022videomae,videomaev2}, MAE-ST~\cite{feichtenhofer2022masked} and Recurrent Video Masked Autoencoders (RVM)~\cite{recurrentvideo}, learn representations by reconstructing missing visual information, while predictive approaches such as V-JEPA~\cite{bardes2024revisiting,vjepa2} learn temporal structure through prediction in feature space. Other approaches explicitly focus on motion, including motion-guided masking~\cite{huang2023mgmae}, motion-trajectory targets~\cite{sun2023mme}, WhatMoves~\cite{fundel2026moves}, which learns localized motion representations, and TIME~\cite{timemachine}, which explores motion-centric representations based on point trajectories. Although these models capture temporal information, their representations are generally learned from videos in which camera and scene motion occur simultaneously. Hence, it is unclear how much the interference between the two sources of motion affects their performance. We investigate this question across video models with different architectures and learning objectives.

\vspace{.5em}\noindent\textbf{Camera Motion Understanding.}
Camera motion has long been estimated geometrically, through structure from motion and
SLAM~\cite{schoenberger2016structure,teed2021droid}. Recent feed-forward models regress pointmaps from which poses
can be recovered~\cite{wang2024dust3r,leroy2024grounding} or predict camera parameters and geometry
directly~\cite{wang2025vggt,vggtomega}, also in dynamic scenes and streaming settings~\cite{zhang2025monst3r,wang2025continuous}.
A complementary line asks whether models understand camera motion semantically. CameraBench~\cite{lin2025camerabench}
annotates videos with a cinematographer-designed taxonomy of camera-motion primitives and compares geometric methods
and VLMs on classification, question answering, captioning and retrieval. CamReasoner~\cite{wu2026camreasoner} trains a
VLM to reason step by step about camera movement, and ACaM~\cite{tan2026natural} tests the recognition of single
cinematic movements on real and synthetic videos. These works study camera motion on its own, whereas we study how it
interferes with scene motion in video representations.

\vspace{.5em}\noindent\textbf{Scene Motion Understanding.}
Many actions are defined by how people and objects move. Third-person benchmarks such as Something-Something
v2~\cite{ssv2} and Diving48~\cite{li2018resound} were built so that classes cannot be told apart by static appearance
alone, while EPIC-KITCHENS~\cite{Damen2022RESCALING} is recorded with head-mounted cameras and
Ego-Exo4D~\cite{grauman2024egoexo4d} pairs head-worn video with synchronized exocentric views of skilled activities.
In all of them, camera and scene motion occur in the same
videos, and camera motion can itself carry information about the action~\cite{xue2026seeing}. We use them as
downstream tasks and complement them with \bench{}.

\vspace{.5em}\noindent\textbf{Camera and Scene Motion in Video Representations.}
Many methods treat the two motions explicitly, by compensating for camera motion in action
recognition~\cite{wang2013action,simonyan2014two}, estimating geometry in dynamic scenes~\cite{zhang2025monst3r} or
controlling both in generation~\cite{wang2024motionctrl}. Knobel~\etal~\cite{knobel2026self} structure temporal
change into a primary component, which captures camera motion when the camera moves, and residual dynamics, learned on
top of frozen image features. While these works design new models, we measure how strongly the two motions interfere in existing video
foundation models and map their fixed features post hoc in each direction.

\providecommand{\bench}{Freeze}%

\section{Do Camera and Scene Motion Interfere with Each Other?}
\label{sec:simcambench}

\subsection{The \bench~Benchmark}
\label{sec:bench}

To study whether and how much camera and scene motion interfere with each other in video representations, we build \bench{}, a
synthetic benchmark in which one source of motion can be frozen while everything else stays fixed.
Since scene motion in real videos ranges from objects driven by physics to moving human bodies, we cover both with two
synthetic subsets: simulated object physics (\emph{\bench-Object}) and animated human characters (\emph{\bench-Human}).
Every clip comes in three versions with identical content and first frame (Fig.~\ref{fig:simcam_viz}): full
(camera and scene move), camera frozen and scene frozen.

\newlength\vizw
\newsavebox\vizrow
\newcommand\vizcols{%
  \makebox[0.19379\vizw]{First frame}\hspace{0.007758\vizw}\makebox[0.19379\vizw]{Last frame}\hspace{0.007758\vizw}%
  \makebox[0.19379\vizw]{Temporal mean}\hspace{0.007758\vizw}\makebox[0.19379\vizw]{Point tracks}\hspace{0.007758\vizw}%
  \makebox[0.19379\vizw]{Motion energy}}
\newcommand\vizrowlab[2]{%
  \sbox\vizrow{\includegraphics[width=\vizw]{fig/simcam_viz_obj_r#2.pdf}}%
  \makebox[0.22in][r]{\raisebox{\dimexpr0.5\ht\vizrow-0.5\height\relax}{\fontsize{6}{6.6}\selectfont\shortstack[r]{#1}}}\hspace{0.03in}%
  \usebox\vizrow\hspace{0.10in}\includegraphics[width=\vizw]{fig/simcam_viz_hum_r#2.pdf}}
\begin{figure*}[t]
  \setlength\vizw{\dimexpr(\textwidth-0.35in)/2\relax}%
  {\fontfamily{LinuxLibertineT-TLF}\selectfont\fontsize{6.5}{7.5}\selectfont\setlength{\lineskip}{0pt}\setlength{\baselineskip}{0pt}%
  \noindent\hspace*{0.25in}\vizcols\hspace{0.10in}\vizcols\par\vspace{3pt}
  \noindent\vizrowlab{Full}{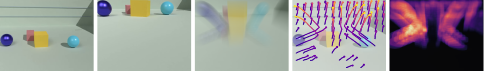}\par\vspace{0.025in}
  \noindent\vizrowlab{Camera\\frozen}{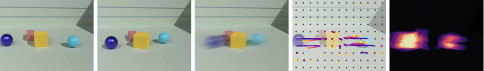}\par\vspace{0.025in}
  \noindent\vizrowlab{Scene\\frozen}{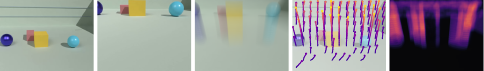}\par}
  \caption{Examples of a \bench-Object clip (left) and a \bench-Human clip (right), each in its three versions. In addition
  to the first and last frames, we present the temporal mean, point tracks and motion energy (mean optical-flow magnitude) to
  show how the motion differs within a triplet.}
  \label{fig:simcam_viz}
\end{figure*}

\vspace{.5em}\noindent\textbf{\bench-Object.}
To build the object subset, we use the synthetic data generator Kubric~\cite{kubric} to render videos in which object
physics and camera motion are set independently.
Each clip places one to five simple rigid shapes in a room and simulates one of four types of physics: collision,
drop/bounce, rolling or static.
Meanwhile, the camera trajectory combines up to seven motion axes (three translations, pan, tilt, roll and zoom).
All trajectories start from a shared pose distribution, so that the first frame reveals little about the camera motion.
Rendering each simulation again with the camera, or every object, held at its first-frame state gives the two frozen
versions.
Finally, following CameraBench's definition~\cite{lin2025camerabench}, we label the camera motion with 15 binary primitives,
computed from the ground-truth camera parameters.
In total, \bench-Object contains 500 triplets in 80 rooms for evaluation.

\begin{table*}[t]
\centering
\small
\setlength{\tabcolsep}{3.2pt}
\begin{tabular}{@{}l >{\centering\arraybackslash}p{24pt}>{\centering\arraybackslash}p{24pt}>{\centering\arraybackslash}p{24pt}@{\hspace{7pt}} >{\centering\arraybackslash}p{24.5pt}>{\centering\arraybackslash}p{24.5pt}>{\centering\arraybackslash}p{24.5pt}@{\hspace{7pt}} >{\centering\arraybackslash}p{30.3pt}>{\centering\arraybackslash}p{30.3pt}>{\centering\arraybackslash}p{30.3pt}@{\hspace{7pt}} >{\centering\arraybackslash}p{32.3pt}>{\centering\arraybackslash}p{32.3pt}>{\centering\arraybackslash}p{32.3pt}@{}}
\toprule
 & \multicolumn{3}{c}{Camera, Object (mAP)} & \multicolumn{3}{c}{Camera, Human (mAP)} & \multicolumn{3}{c}{Physics family, Object (acc.)} & \multicolumn{3}{c}{Same moment, Human (AUC)} \\
\cmidrule(l{9pt}r{7pt}){2-4}\cmidrule(l{6pt}r{7pt}){5-7}\cmidrule(l{8.9pt}r{7pt}){8-10}\cmidrule(l{9.9pt}r{0pt}){11-13}
Backbone & full & {\renewcommand{\arraystretch}{0.72}\begin{tabular}{@{}c@{}}scene\\frozen\end{tabular}} & $\Delta$ & full & {\renewcommand{\arraystretch}{0.72}\begin{tabular}{@{}c@{}}scene\\frozen\end{tabular}} & $\Delta$ & full & {\renewcommand{\arraystretch}{0.72}\begin{tabular}{@{}c@{}}camera\\frozen\end{tabular}} & $\Delta$ & full & {\renewcommand{\arraystretch}{0.72}\begin{tabular}{@{}c@{}}camera\\frozen\end{tabular}} & $\Delta$ \\
\midrule
V-JEPA 2.1~\cite{vjepa21} & 71.1 & 76.6 & \textcolor{gaincolor}{$+5.5$} & 83.0 & 86.1 & \textcolor{gaincolor}{$+3.1$} & 69.4 & 88.2 & \textcolor{gaincolor}{$+18.8$} & 53.1 & 67.4 & \textcolor{gaincolor}{$+14.3$} \\
VideoMAEv2~\cite{videomaev2} & 42.6 & 46.3 & \textcolor{gaincolor}{$+3.7$} & 48.9 & 57.1 & \textcolor{gaincolor}{$+8.3$} & 88.0 & 92.8 & \textcolor{gaincolor}{$+4.8$} & 65.6 & 74.2 & \textcolor{gaincolor}{$+8.6$} \\
RVM~\cite{recurrentvideo} & 66.5 & 76.6 & \textcolor{gaincolor}{$+10.1$} & 94.3 & 95.2 & \textcolor{gaincolor}{$+0.9$} & 74.0 & 86.0 & \textcolor{gaincolor}{$+12.0$} & 55.3 & 66.6 & \textcolor{gaincolor}{$+11.3$} \\
WhatMoves~\cite{fundel2026moves} & 76.1 & 88.2 & \textcolor{gaincolor}{$+12.1$} & 91.4 & 94.6 & \textcolor{gaincolor}{$+3.2$} & 83.0 & 91.0 & \textcolor{gaincolor}{$+8.0$} & 70.8 & 89.4 & \textcolor{gaincolor}{$+18.7$} \\
\bottomrule
\end{tabular}
\caption{For all four backbones, probes score higher on every task when the other motion is frozen. All scores are in \%.
$\Delta$ is the frozen score minus the full score in points. Their confidence intervals are given in Table~\ref{tab:app-main-ci}.}
\label{tab:simcambench-results}
\end{table*}

\vspace{.5em}\noindent\textbf{\bench-Human.}
To build the human subset, we turn to an existing dataset called Cam$\times$Time~\cite{huang2025spacetimepilot} that provides disentangled camera and scene motion.
Each scene films an animated human character along four camera trajectories, and every moment of the motion is
rendered from every camera position along a trajectory. In a full clip, the camera moves along the trajectory as time
runs on. Holding the camera at its first position gives the camera-frozen version, and holding the motion at its first
moment gives the scene-frozen version.
We build the subset from all 417 scenes of the release available to us, cutting one clip per trajectory in its three
versions.
We label these clips with the same primitives as for \bench-Object, except zoom and roll, which are absent or
predictable from the first frame in this dataset.
Appendix~\ref{app:simcambench} describes how both subsets are created and labeled.

\subsection{Evaluating Video Representations on \bench}
\label{sec:bench-eval}

\noindent\textbf{Tasks.}
To measure the interference in both directions, we need a task for each kind of motion.
For the \emph{camera} task, we predict the motion primitives on both subsets, with one binary probe per primitive.
We score it by mean average precision (mAP) and compare full clips with their scene-frozen versions.
For scene motion, the camera-frozen versions serve as the reference, and each subset has its own task.
On \bench-Object, we classify the \emph{physics family} of each clip and report accuracy.
The human clips have no such labels, so on \bench-Human we instead ask whether two windows of one scene show the
\emph{same moment}, even though they were filmed along different trajectories.
Answering this requires matching the person's motion across viewpoints.
All trajectories of a scene film the character's motion over the same time steps.
We therefore cut each of them into the same three consecutive 40-step windows, and two windows from different
trajectories show the same moment exactly when they cover the same steps.
For each such pair, a logistic regression takes the absolute difference of the two window features as input and predicts whether
they show the same moment.
We report ROC-AUC over all pairs, for full and camera-frozen windows alike.
Further details on the tasks and the feature extraction are given in Appendix~\ref{app:simcambench}.

\vspace{.5em}\noindent\textbf{Evaluation.}
We evaluate four recent pre-trained video backbones: \mbox{V-JEPA 2.1}~\cite{vjepa21}, VideoMAEv2~\cite{videomaev2},
RVM~\cite{recurrentvideo} and WhatMoves~\cite{fundel2026moves}.
To measure what their representations already encode, we keep every backbone fixed and evaluate it only with linear probes.
Each backbone sees 16 frames sampled uniformly over a clip, and we average its 8 temporal step features into one clip
feature.
For each task, we train one probe on the full clips and another on their frozen versions.
Since the benchmark has no fixed split, we use five-fold cross-validation: each probe is trained on the clips of four
fifths of the rooms (or scenes) and tested on the rest, so every clip is scored once and no room or scene appears in
both training and test. Both probes use the same folds.
The gap between the two scores thus shows how much easier the task becomes once the other motion is frozen.
As Table~\ref{tab:simcambench-results} shows, every backbone does better on every task once the other motion is frozen.
The interference between camera and scene motion is thus not only intuitive but measurable, and it occurs in both directions.
This suggests that considering the two separately could help, and that current pre-trained features leave clear room for
improvement.

\providecommand{\cam}{CamBit}
\providecommand{\scene}{SceneBit}
\section{Method}
\label{sec:method}

\begin{figure}[t]
\centering
\includegraphics[width=\linewidth]{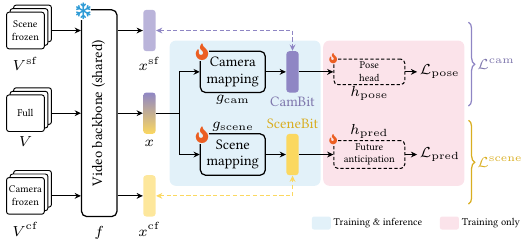}
\caption{Overview of the model that extracts \cam{} and \scene{} from video features from a given backbone. To train, we use triplets of nearly-identical videos, where either: 1) both scene and camera move (``Full"), 2) only the camera moves (``Scene frozen") or 3) only the scene moves (``Camera frozen"). We train two small networks (``Camera mapping" and ``Scene mapping") to map the original features to each of the two other features in the triplet. We use auxiliary tasks (Recovering pose head and future anticipation) to further constrain the optimization when using real-world videos.}
\label{fig:overview}
\vspace{-1pt}  %
\end{figure}

\noindent\textbf{Overview.}
Sec.~\ref{sec:simcambench} shows that video representations capture each kind of motion better once the other is
frozen.
This raises the question of whether we can learn to map the features of the full clip to those of a frozen version.
If so, we can apply the mapping post hoc to real videos, without re-training the backbone.

Given a backbone $f$, we encode a clip $V$ as features $x = f(V) \in \mathbb{R}^{T \times D}$. For each of its $T{=}8$
temporal steps, we average the last-layer patch tokens over space into one $D$-dimensional vector, or take the global
motion token of that step for WhatMoves.
\bench{} provides the training pairs: for every synthetic clip, we denote the features of its scene-frozen and
camera-frozen versions by $x^{\mathrm{sf}}$ and $x^{\mathrm{cf}}$.
We learn one mapping towards $x^{\mathrm{sf}}$, which gives the camera representation \cam, and one towards
$x^{\mathrm{cf}}$, which gives the scene representation \scene{} (Fig.~\ref{fig:overview}).
The two mappings share one small architecture and return features of the same shape as $x$, so either representation can
be used wherever $x$ is, for example as the input of a linear probe.
Since frozen versions exist only for synthetic clips, we also train each mapping with signals from real videos.
Sec.~\ref{sec:experiments} evaluates both mappings on real videos, and Sec.~\ref{sec:exp-analysis} analyzes what makes
each direction easy or hard from several aspects, such as the role of camera motion and feature space distances.

\subsection{Camera Mapping}
\label{sec:method-camera}

The camera mapping $g_{\mathrm{cam}}: \mathbb{R}^{T \times D} \rightarrow \mathbb{R}^{T \times D}$ is designed to bring
$x$ closer to the features of its scene-frozen version, with clearer camera information (Sec.~\ref{sec:simcambench}).
Since $x$ already encodes camera motion (Table~\ref{tab:simcambench-results}), we let a network predict only a
correction to $x$, which is easier to learn than the full output~\cite{he2016deep}:
\begin{equation}\label{eq:main-1}
\cam = g_{\mathrm{cam}}(x) = x - w_\psi(x).
\end{equation}
Here $w_\psi$ is a small Transformer encoder~\cite{vaswani2017attention} that takes the $T$ steps as tokens.
Its configuration and the computation of the pose targets are detailed in Appendix~\ref{app:method}.

\vspace{.5em}\noindent\textbf{Learning from \bench.}
\bench{} contains the scene-frozen version of every clip, and its features $x^{\mathrm{sf}}$ serve as the target:
\begin{equation}\label{eq:main-2}
\mathcal{L}^{\mathrm{cam}}_{\mathrm{syn}} = \| g_{\mathrm{cam}}(x) - x^{\mathrm{sf}} \|^2
  + \| g_{\mathrm{cam}}(x^{\mathrm{sf}}) - x^{\mathrm{sf}} \|^2,
\end{equation}
where $\|\cdot\|^2$ is the squared error averaged over steps and dimensions.
The second term asks the mapping to leave a clip unchanged when its scene is already frozen.

\vspace{.5em}\noindent\textbf{Learning from real videos.}
Since the mapping needs to work on real videos, training it only on rendered clips leaves a domain gap.
To close this gap, we supervise \cam{} with the camera motion of real clips.
We use clips from Condensed Movies~\cite{bain2020condensed} with camera poses estimated by
\mbox{VGGT-$\Omega$}~\cite{vggtomega}.
To reduce the effect of noise in the estimated poses, we also include clips from BEDLAM2~\cite{tesch2026bedlam2}, a
photorealistic synthetic dataset with ground-truth camera poses.
Following~\cite{zhou2026unveiling}, we describe the camera motion from step $t$ to $t{+}1$ by a unit direction
$d^{m}_t$ and a log speed $s^{m}_t$, once for the translation ($m = \mathrm{tr}$) and once for the rotation
($m = \mathrm{rot}$).
A small Transformer head $h_{\mathrm{pose}}$ takes all $T$ steps of \cam{} and predicts them for every pair of
consecutive steps, $\{(\hat d_t, \hat s_t)\}_{t=1}^{T-1} = h_{\mathrm{pose}}(\cam)$; it compresses its features into a
compact code before a final linear layer. It is trained with
\begin{equation}\label{eq:main-3}
\begin{aligned}
\mathcal{L}_{\mathrm{pose}} = \sum_{m \in \{\mathrm{tr},\,\mathrm{rot}\}} \Big[
  & \frac{1}{|\mathcal{S}^{m}|} \sum_{t \in \mathcal{S}^{m}} \big( 1 - \cos(\hat d^{m}_t, d^{m}_t) \big) \\
  & + \frac{1}{T-1} \sum_{t=1}^{T-1} \ell_{\mathrm{H}}\big(\hat s^{m}_t - s^{m}_t\big) \Big],
\end{aligned}
\end{equation}
where $\ell_{\mathrm{H}}$ is the Huber loss. The direction terms are averaged only over the steps $\mathcal{S}^{m}$ in
which the camera translates or rotates, because a camera at rest has no direction of motion.

Therefore, the full objective of the camera mapping is
\begin{equation}\label{eq:main-4}
\mathcal{L}^{\mathrm{cam}} = \mathbb{E}_{\mathcal{D}_{\mathrm{syn}}}\big[\mathcal{L}^{\mathrm{cam}}_{\mathrm{syn}}\big]
  + \mathbb{E}_{\mathcal{D}_{\mathrm{pose}}}\big[\mathcal{L}_{\mathrm{pose}}\big],
\end{equation}
where $\mathcal{D}_{\mathrm{syn}}$ denotes the \bench{} triplets and $\mathcal{D}_{\mathrm{pose}}$ the real and BEDLAM2 clips, each
sampled in its own batch. The head $h_{\mathrm{pose}}$ is used only during training.

\subsection{Scene Mapping}
\label{sec:method-scene}

The scene mapping $g_{\mathrm{scene}}$ aims to bring $x$ closer to the features of its camera-frozen version, to get a
cleaner representation of the scene motion. It has the same form as $g_{\mathrm{cam}}$, with a network $u_\phi$ of the same architecture as
$w_\psi$:
\begin{equation}\label{eq:main-5}
\scene = g_{\mathrm{scene}}(x) = x - u_\phi(x).
\end{equation}

\vspace{.5em}\noindent\textbf{Learning from \bench.}
We train $g_{\mathrm{scene}}$ on \bench{} in the same way as $g_{\mathrm{cam}}$, with the camera-frozen version as the
target:
\begin{equation}\label{eq:main-6}
\mathcal{L}^{\mathrm{scene}}_{\mathrm{syn}} = \| g_{\mathrm{scene}}(x) - x^{\mathrm{cf}} \|^2
  + \| g_{\mathrm{scene}}(x^{\mathrm{cf}}) - x^{\mathrm{cf}} \|^2 .
\end{equation}

\vspace{.5em}\noindent\textbf{Learning from real videos.}
As for the camera mapping, we also train on real videos to reduce the domain gap. For this direction, we approximate
real pairs through augmentation.
On clips filmed with a still camera, we replay a real camera path, a sequence of rotations and zooms, as a moving crop
$\tilde x$, and take a fixed crop $\bar x$ of the same clip as its camera-frozen counterpart.
We train the mapping to turn the first into the second and to leave the second unchanged:
\begin{equation}\label{eq:main-7}
\mathcal{L}_{\mathrm{crop}} = \| g_{\mathrm{scene}}(\tilde x) - \bar x \|^2
  + \| g_{\mathrm{scene}}(\bar x) - \bar x \|^2 .
\end{equation}
The squared error penalizes remaining camera motion only on average over dimensions.
To constrain it more directly, we encourage \scene{} to be invariant to camera motion with a symmetric InfoNCE
loss~\cite{oord2018cpc} on clip-level embeddings, which a small embedding network $\varphi$ computes from each
representation, following common practice in contrastive learning~\cite{chen2020simple}. It pulls together the two versions of a clip, which differ only in camera motion, and pushes apart
different clips:
\begin{equation}\label{eq:main-8}
\begin{aligned}
\mathcal{L}_{\mathrm{inv}} = \tfrac{1}{2}\big[\, & \mathrm{NCE}\big(\varphi(g_{\mathrm{scene}}(x)), \varphi(x^{\mathrm{cf}})\big) \\
  & + \mathrm{NCE}\big(\varphi(g_{\mathrm{scene}}(\tilde x)), \varphi(g_{\mathrm{scene}}(\bar x))\big) \big].
\end{aligned}
\end{equation}

\vspace{.5em}\noindent\textbf{Keeping scene motion.}
The invariance term, however, would be met just as well by a representation that carries no motion at all, discarding
the scene motion along with the camera motion.
To counter this, we introduce a small recurrent prediction head $h_{\mathrm{pred}}$ to preserve the scene motion.
$h_{\mathrm{pred}}$ predicts the next step $x_{t+1}$ of the original features of a real clip from the current and previous steps of
\scene{}:
\begin{equation}\label{eq:main-9}
\mathcal{L}_{\mathrm{pred}} = \frac{1}{T-1} \sum_{t=1}^{T-1} \big\| h_{\mathrm{pred}}(\scene_{1:t}) - x_{t+1} \big\|^2 .
\end{equation}
We predict $x$ rather than \scene{} itself, because a motion-free \scene{} would make its own future trivially
predictable, while $x$ keeps the actual dynamics of the clip.

Therefore, the full objective of the scene mapping is
\begin{equation}\label{eq:main-10}
\begin{aligned}
\mathcal{L}^{\mathrm{scene}} = {} & \mathbb{E}_{\mathcal{D}_{\mathrm{syn}}}\big[\mathcal{L}^{\mathrm{scene}}_{\mathrm{syn}}\big]
  + \mathbb{E}_{\mathcal{D}_{\mathrm{crop}}}\big[\mathcal{L}_{\mathrm{crop}}\big] \\
  & + \lambda_{\mathrm{inv}}\,\mathbb{E}_{\mathcal{D}_{\mathrm{syn}} \cup \mathcal{D}_{\mathrm{crop}}}\big[\mathcal{L}_{\mathrm{inv}}\big]
  + \lambda_{\mathrm{pred}}\,\mathbb{E}_{\mathcal{D}_{\mathrm{pred}}}\big[\mathcal{L}_{\mathrm{pred}}\big],
\end{aligned}
\end{equation}
where $\mathcal{D}_{\mathrm{crop}}$ and $\mathcal{D}_{\mathrm{pred}}$ denote the crop pairs and the real clips of the terms above,
each sampled in its own batch, and the weights $\lambda$ are set once before training to balance the initial scales of
the terms. The embedding network $\varphi$ and the prediction head $h_{\mathrm{pred}}$ are used only during
training. We list the training sets, schedules and loss weights of both mappings in Appendix~\ref{app:method}.

\section{Experiments}
\label{sec:experiments}

\subsection{Experimental Settings}
\label{sec:exp-setting}

The proposed \cam{} and \scene{} are trained to bring the representation of a video clip closer to that of its frozen versions
(Sec.~\ref{sec:simcambench}), so that it contains cleaner information on each kind of motion and performs better on
downstream tasks. We now evaluate these representations on real videos and on held-out synthetic scenes.

\vspace{.5em}\noindent\textbf{Training data.} We use a combination of synthetic and real-world data to train the \scene{} and \cam{} models. 

\noindent\emph{Synthetic data:}
To train the mappings, we extend \bench{} with the same pipelines to 11,990
Cam$\times$Time clips from 300 scenes and 800 \bench-Object triplets, partly rendered from simulations not used in
Sec.~\ref{sec:simcambench}. We keep 84 Cam$\times$Time scenes and 12 \bench-Object rooms out of all training and
model selection, and evaluate the mappings only on these held-out \bench{} scenes.

\noindent\emph{Real-world data:} Besides these pairs, the camera mapping trains on 47,636 real-world clips from Condensed Movies and 24,709 clips from
BEDLAM2, while the scene mapping builds its augmented pairs from 4,799 clips of Something-Something v2
(SSv2)~\cite{ssv2} filmed with a nearly still camera and uses
5,700 further SSv2 clips, drawn regardless of their camera motion, for self-supervised next-step prediction.
How these training sets are built is explained in Appendix~\ref{app:method-data}.

\vspace{.5em}\noindent\textbf{Evaluation.} We evaluate using the features from the four different backbones we have used before in Sec.~\ref{sec:simcambench}. These original features are input to our model to obtain a \scene{} and a \cam{} for each of them. We then use linear probing to compare the original and the transformed \scene{} and \cam{} in a variety of scene tasks and camera tasks. We report the difference averaged over three training seeds.
Appendix~\ref{app:exp} reports the protocol of every task, and Appendix~\ref{app:ablations} ablations of the training terms of both mappings.

\subsection{Tasks}
\label{sec:exp-tasks}

\noindent\textbf{Camera understanding.}
We use CameraBench~\cite{lin2025camerabench}, whose 15 camera-motion primitives, such as pans and dollies, follow how
cinematographers describe camera movement.
For each primitive, a logistic-regression probe is trained on the official training videos, and we report the mean
average precision (mAP) over the primitives on the test videos.
Since camera movement is often described in words, we also evaluate CameraBench's skill-based retrieval, which matches
videos with questions about their camera motion. Here a linear head maps the representation into a frozen text
space~\cite{zhu2024languagebind}, and we report CameraBench's group score, which counts an item only if both the
video-to-text and the text-to-video matches are correct.
We also include ACaM~\cite{tan2026natural}, a more recent benchmark of camera movements described in natural language.
Following its four-option protocol, a probe trained on the ACaM training set picks one of four movements for each
test video, and we report the accuracy averaged over the 17 movements.
We also classify the camera motion of these \bench{} scenes, following the protocol of
Sec.~\ref{sec:simcambench}.

\vspace{.5em}\noindent\textbf{Geometric grounding.}
To test whether the representation is able to track the camera's pose over time, a linear regressor is employed to
predict the relative 6-DoF pose of the camera between two steps.
Based on our observation in Sec.~\ref{sec:simcambench} that scene motion can interfere with camera information, the goal is to measure how the camera mapping performs under different amounts of scene motion. We use real
videos from DL3DV~\cite{ling2024dl3dv} with static scenes, EMDB~\cite{kaufmann2023emdb} with moving people, and the
dynamic sequences of TUM RGB-D~\cite{sturm12iros}, and report the mean squared error of the standardized pose (zMSE)
on each.

\vspace{.5em}\noindent\textbf{Scene understanding.}
For the scene experiments, we use tasks that depend on how people and objects move.
We use two standard benchmarks of motion-centric action recognition, Diving48~\cite{li2018resound}
and SSv2~\cite{ssv2}.
We also evaluate on egocentric videos, a different domain in which the head-mounted camera tends to move more
than in third-person footage. %
We use EPIC-KITCHENS-100~\cite{Damen2022RESCALING} and Ego-Exo4D~\cite{grauman2024egoexo4d}. In these datasets the head motion varies widely, hence for training we only use clips with little head motion. We evaluate on all validation clips (All) and on those with much head motion (High), to observe
how performance varies as a function of camera motion. 
Finally, we also evaluate on the two scene tasks of Sec.~\ref{sec:simcambench} on the \bench{} scenes. 
We report class-balanced accuracy to address the long-tailed class distribution.

\subsection{Camera Results}
\label{sec:exp-camera}

\providecommand{\dlt}[2]{\,{\scriptsize$#1{\pm}#2$}}
\providecommand{\dn}[1]{\,{\scriptsize$#1$}}
\providecommand{\best}[1]{\cellcolor{bestcell}\textbf{#1}}
\providecommand{\gain}[1]{\,{\scriptsize\textcolor{gaincolor}{$#1$}}}
\providecommand{\drop}[1]{\,{\scriptsize\textcolor{losscolor}{$#1$}}}

Table~\ref{tab:camera-real} compares \cam{} with original  features from four backbones.
\cam{} improves on nearly every task and backbone.
We observe that \cam{} improves the performance across datasets and backbones, sometimes up to close to 20\%. These results suggest that indeed there is value in recovering clearer camera information from mixed representations, as standard foundation models struggle to do so by default. These results also show that it is not necessary to fully re-train the original foundation models from scratch, and that instead the camera information is implicit and can be extracted in a post-hoc manner.

\begin{table*}[ht!]
\centering
\definecolor{maprow}{HTML}{E7EEF9}
\small
\setlength{\tabcolsep}{2.6pt}
\begin{tabular}{@{}l ccccc ccc@{}}
\toprule
 & \multicolumn{5}{c}{Camera understanding $\uparrow$} & \multicolumn{3}{c}{Geometric grounding $\downarrow$} \\
\cmidrule(lr){2-6}\cmidrule(lr){7-9}
 & \multicolumn{2}{c}{CameraBench~\cite{lin2025camerabench}} & ACaM~\cite{tan2026natural} & \bench-O. & \bench-H.
 & DL3DV~\cite{ling2024dl3dv} & EMDB~\cite{kaufmann2023emdb} & TUM RGB-D~\cite{sturm12iros} \\
\cmidrule(lr){2-3}
 & mAP & Retrieval & Acc. & mAP & mAP & zMSE & zMSE & zMSE \\
\midrule
V-JEPA 2.1~\cite{vjepa21}  & 70.4 & 31.0 & 74.1 & 53.8 & 74.8 & .721 & .503 & .879 \\
\rowcolor{maprow}\quad + \cam  & \textbf{84.0}\gain{+13.6} & \textbf{41.5}\gain{+10.5} & \textbf{81.8}\gain{+7.7} & \textbf{60.3}\gain{+6.5} & \textbf{93.8}\gain{+19.0} & \textbf{.643}\gain{-.078} & \textbf{.200}\gain{-.303} & \textbf{.421}\gain{-.458} \\
\midrule
WhatMoves~\cite{fundel2026moves}  & 70.8 & 32.9 & 73.2 & 60.1 & 92.2 & .595 & .335 & .420 \\
\rowcolor{maprow}\quad + \cam  & \textbf{81.3}\gain{+10.5} & \textbf{41.2}\gain{+8.3} & \textbf{77.9}\gain{+4.7} & \textbf{66.1}\gain{+6.0} & \textbf{92.6}\gain{+0.4} & \textbf{.586}\gain{-.009} & \textbf{.249}\gain{-.086} & \textbf{.345}\gain{-.075} \\
\midrule
VideoMAEv2~\cite{videomaev2}  & 50.7 & 20.9 & 61.9 & 29.1 & 40.0 & .884 & .778 & 1.026 \\
\rowcolor{maprow}\quad + \cam  & \textbf{70.4}\gain{+19.7} & \textbf{39.3}\gain{+18.4} & \textbf{71.2}\gain{+9.3} & \textbf{44.6}\gain{+15.5} & \textbf{80.7}\gain{+40.7} & \textbf{.771}\gain{-.113} & \textbf{.445}\gain{-.333} & \textbf{.794}\gain{-.232} \\
\midrule
RVM~\cite{recurrentvideo}  & 70.4 & 29.9 & 69.7 & 50.4 & 87.3 & .704 & .405 & .700 \\
\rowcolor{maprow}\quad + \cam  & \textbf{82.2}\gain{+11.8} & \textbf{41.5}\gain{+11.6} & \textbf{79.2}\gain{+9.5} & \textbf{64.6}\gain{+14.2} & \textbf{94.1}\gain{+6.8} & \textbf{.656}\gain{-.048} & \textbf{.303}\gain{-.102} & \textbf{.453}\gain{-.247} \\
\bottomrule
\end{tabular}
\caption{Camera results. Each blue \cam{} row reads \cam{} computed from the backbone features of the row above, with
the same probe. \bench-O.\ and \bench-H.: the held-out scenes of \bench-Object and \bench-Human. Understanding scores are
in \%, and zMSE is the mean squared error of the standardized pose (lower is better). Retrieval: CameraBench's skill-based
retrieval (group score). Green numbers give the change from the backbone row, and the better value of each pair is in bold.}
\label{tab:camera-real}
\end{table*}

\subsection{Scene Results}
\label{sec:exp-scene}

Table~\ref{tab:scene-real} compares \scene{} with the original backbone features.
On the \bench{} scenes, \scene{} improves physics-family accuracy on all four backbones and same-moment AUC on
three of them.
On real videos, the effects are more nuanced. We test across a wide variety of datasets, both exo-centric and ego-centric, with varying degrees of camera motion, and diverse domains. We observe that the gains are overall moderate, which motivates us to analyze and explore in depth the reasons behind this effect in Sec.~\ref{sec:exp-analysis}.

\begin{table*}[t]
\centering
\definecolor{maprow}{HTML}{E7EEF9}
\small
\setlength{\tabcolsep}{4pt}
\begin{tabular}{@{}l cccccccc@{}}
\toprule
 & \bench-O. & \bench-H. & Diving48~\cite{li2018resound} & SSv2~\cite{ssv2} & \multicolumn{2}{c}{EPIC-K.~\cite{Damen2022RESCALING}} & \multicolumn{2}{c}{Ego-Exo4D~\cite{grauman2024egoexo4d}} \\
\cmidrule(lr){6-7}\cmidrule(lr){8-9}
 & Acc. & AUC & Acc. & Acc. & Acc.\ (All) & Acc.\ (High) & Acc.\ (All) & Acc.\ (High) \\
\midrule
V-JEPA 2.1~\cite{vjepa21}  & 23.4 & 51.6 & 25.7 & 39.8 & 33.8 & 24.5 & 16.4 & 12.2 \\
\rowcolor{maprow}\quad + \scene  & \textbf{35.3}\gain{+11.9} & \textbf{52.7}\gain{+1.1} & \textbf{26.0}\gain{+0.3} & \textbf{39.9}\gain{+0.1} & \textbf{34.7}\gain{+0.9} & \textbf{25.5}\gain{+1.0} & \textbf{17.0}\gain{+0.6} & \textbf{12.5}\gain{+0.3} \\
\midrule
WhatMoves~\cite{fundel2026moves}  & 31.3 & 61.6 & 22.0 & 24.4 & 24.8 & 18.4 & 11.3 & 8.3 \\
\rowcolor{maprow}\quad + \scene  & \textbf{51.7}\gain{+20.4} & \textbf{70.1}\gain{+8.5} & \textbf{22.3}\gain{+0.3} & \textbf{24.5}\gain{+0.1} & \textbf{25.3}\gain{+0.5} & \textbf{18.9}\gain{+0.5} & \textbf{11.8}\gain{+0.5} & \textbf{8.7}\gain{+0.4} \\
\midrule
VideoMAEv2~\cite{videomaev2}  & 51.7 & \textbf{57.9} & 27.2 & \textbf{30.6} & \textbf{30.3} & 22.4 & \textbf{20.1} & \textbf{16.7} \\
\rowcolor{maprow}\quad + \scene  & \textbf{57.7}\gain{+6.0} & 57.6\drop{-0.3} & \textbf{27.3}\gain{+0.1} & \textbf{30.6}\gain{+0.0} & \textbf{30.3}\gain{+0.0} & \textbf{23.0}\gain{+0.6} & 20.0\drop{-0.1} & \textbf{16.7}\gain{+0.0} \\
\midrule
RVM~\cite{recurrentvideo}  & 18.9 & 53.4 & 20.1 & 36.8 & 29.9 & 20.5 & 13.4 & 10.1 \\
\rowcolor{maprow}\quad + \scene  & \textbf{30.8}\gain{+11.9} & \textbf{55.3}\gain{+1.9} & \textbf{21.2}\gain{+1.1} & \textbf{36.9}\gain{+0.1} & \textbf{30.2}\gain{+0.3} & \textbf{21.5}\gain{+1.0} & \textbf{14.3}\gain{+0.9} & \textbf{11.8}\gain{+1.7} \\
\bottomrule
\end{tabular}
\caption{Scene results. Each blue \scene{} row reads \scene{} computed from the backbone features of the row above, with
the same probe. \bench-O.\ and \bench-H.: physics-family accuracy and same-moment AUC on the held-out scenes. EPIC-K.\ (EPIC-KITCHENS) and Ego-Exo4D: class-balanced accuracy
on all validation clips (All) and on those with much head motion (High). All scores are in \% (higher is better). Green and red
numbers give the increase and decrease from the backbone, and the better value of each pair is in bold.}
\label{tab:scene-real}
\end{table*}

\subsection{Further Analysis}
\label{sec:exp-analysis}

\begin{figure}[t]
\centering
\includegraphics[width=\linewidth]{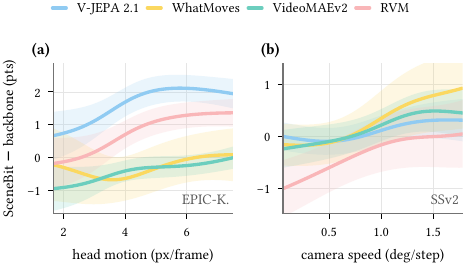}
\caption{\scene{} is more effective in clips with higher camera speed. (a) EPIC-KITCHENS, measured head motion. (b) SSv2, camera speed
estimated with VGGT-$\Omega$. The bands show $\pm$1 standard error of the smoothed gain.}
\label{fig:analysis-speed}
\end{figure}

\noindent\textbf{Q: Does camera speed affect the gain of the scene mapping?}
Many clips are filmed with a nearly still
camera or show a static scene. Such clips contain little camera motion that interferes with the scene, or no scene motion to
recover, so the mapping can add little to what the backbone features already contain.
To see under which conditions the scene mapping helps, we plot the Gaussian-smoothed gain of \scene{} as a function of camera speed in two standard benchmarks: EPIC-KITCHENS and SSv2 (Fig.~\ref{fig:analysis-speed}). We find that indeed there is a correlation between the gain of the scene mapping and the camera speed. This trend is consistent across backbones.

\begin{figure}[t]
\centering
\includegraphics[width=\linewidth]{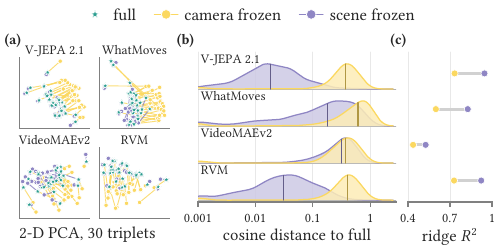}
\caption{The full clip is closer to its scene-frozen version than to its camera-frozen version in feature space (\bench-Human). (a) 2-D PCA of 30 random triplets per backbone, with each full clip joined to its two frozen versions. (b) Cosine distance to the full clip over all test triplets (log scale, vertical lines: medians). (c) $R^2$ of a ridge map from the full clip.}
\label{fig:analysis-geometry}
\end{figure}

\vspace{.5em}\noindent\textbf{Q: Is learning the scene mapping harder than learning the camera mapping? } The camera mapping (Tab.~\ref{tab:camera-real}) shows much clearer gains than the scene mapping (Tab.~\ref{tab:scene-real}), while having a similar architecture and the same synthetic training data. We investigate the reason for this discrepancy by examining the space of triplets of features in the \bench-human benchmark. Results are shown in Fig.~\ref{fig:analysis-geometry}. We observe that features of original video are closer to features of frozen scene videos than to features of camera frozen videos for all four backbones, most clearly for V-JEPA 2.1 and RVM. We explore this in three different ways: visually through a PCA scatterplot, numerically through the distribution of cosine distances, and through the $R^2$ of a ridge regression from the full clip to each frozen version. All three show that the camera-frozen version is farther from the full clip and harder to predict, which suggests that mapping towards the camera-frozen version is indeed harder than mapping towards the scene-frozen version. Intuitively this makes sense, as the visual impact of moving the camera is simply larger, than the effect of moving an object. This suggests that future work might explore this problem beyond the post hoc setting, addressing camera interference already at training time.

\begin{figure}[t]
\centering
\includegraphics[width=\linewidth]{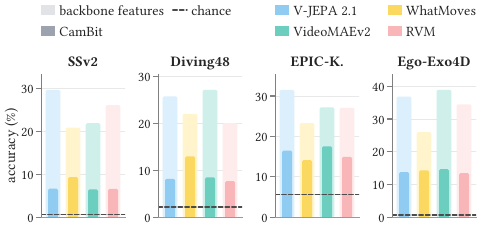}
\caption{Action recognition from \cam{} alone, compared with the backbone features, on four real datasets.}
\label{fig:analysis-camera-only}
\end{figure}

\vspace{.5em}\noindent\textbf{Q: How much information does camera motion contain about actions?}
By design, the scene mapping moves the representation towards a version with less interference from camera motion. This helps when the label is carried by the scene motion alone, the case that \bench{}
isolates and where \scene{} gains clearly. Real action labels, however, are only partly of this kind, since camera motion
can itself carry information about the action~\cite{xue2026seeing}. To see to what extent this holds in our setting, we train a linear probe on
\cam{} alone for each real dataset and compare it with the same probe on the backbone features
(Fig.~\ref{fig:analysis-camera-only}).
On every dataset and backbone, \cam{} alone recognizes actions far above chance.
With less camera information, the mapped representation can therefore also lose part of what these tasks rely on,
which may be one of the reasons for the small gains of \scene{} on real videos.

\vspace{.5em}\noindent\textbf{Q: Why do the gains differ between backbones?}
The gains of both mappings vary considerably between backbones. For example, we observe that VideoMAEv2, whose
features encode camera motion least clearly (Tables~\ref{tab:simcambench-results} and~\ref{tab:camera-real}), gains
the most from the camera mapping, but the least from the scene mapping. Intuitively this makes sense, as a backbone
that represents camera motion only weakly leaves more room to make it explicit, but also less camera interference to
remove. This suggests that the benefit of each mapping depends on how well the backbone encodes the two motions in the
first place.

\vspace{.5em}\noindent\textbf{Q: Which actions benefit from the scene mapping?}
We also group the classes of SSv2 by what their labels describe and compare their accuracy before and after the scene
mapping. On average across backbones, the accuracy improves for classes where an object moves relative to the camera,
suggesting that the scene mapping helps distinguish object motion from the camera's position and motion, while it
drops the most for classes defined by camera motion, showing that \scene{} indeed moves away from camera information. See Appendix~\ref{app:exp-analysis} for per-category results and a more detailed discussion.

\section{Conclusion}
\label{sec:conclusion}

In this paper, we examine the interference between camera motion and scene motion in the representations of video
foundation models.
With \bench{}, we turn it into a controlled measurement and show that it appears consistently across tasks and
backbones.
Our post-hoc mappings further show that this interference can be mitigated without re-training the backbone.
On the camera side, a lightweight mapping on fixed features makes camera information much easier to read on real videos.
On the scene side, the mapping also improves over the backbone features, and our analysis shows that the problem is
more complex on real videos. The scene mapping is harder to learn, camera motion can itself carry
information about the action, and the effect of the mapping varies with the backbone and how strongly the
camera moves.
These findings suggest that video representations can benefit from adapting more robustly to the context-dependent
relationship between camera and scene motion. 
To sum up, \bench{} enables the first controlled measurement of the interference between camera and scene motion, and our mappings provide a simple way to mitigate it on top of various pre-trained video models.
We hope that our work can serve as a basis for future research in this direction and more
broadly for video representation learning.

\section*{LLM Usage Statement}

We used Claude (Anthropic), a large language model, in three parts of this work. For the code, it helped write and debug
the scripts for data generation, training and evaluation. For the figures, it helped write the plotting scripts. For the
writing, it edited the text of all sections for language and clarity. The research questions, experiments, results and
claims were decided and verified by the authors, who take full responsibility for the content of the paper.

{
    \small
    \bibliographystyle{ieeenat_fullname}
    \bibliography{main}

@article{Damen2022RESCALING,
  title   = {Rescaling Egocentric Vision: Collection, Pipeline and Challenges for {EPIC-KITCHENS-100}},
  author  = {Damen, Dima and Doughty, Hazel and Farinella, Giovanni Maria and Furnari, Antonino and Kazakos, Evangelos and Ma, Jian and Moltisanti, Davide and Munro, Jonathan and Perrett, Toby and Price, Will and Wray, Michael},
  journal = {International Journal of Computer Vision (IJCV)},
  volume  = {130},
  pages   = {33--55},
  year    = {2022}
}

@inproceedings{bain2020condensed,
  title     = {Condensed Movies: Story Based Retrieval with Contextual Embeddings},
  author    = {Bain, Max and Nagrani, Arsha and Brown, Andrew and Zisserman, Andrew},
  booktitle = {Proceedings of the Asian Conference on Computer Vision (ACCV)},
  year      = {2020}
}

@article{bardes2024revisiting,
  title={Revisiting Feature Prediction for Learning Visual Representations from Video},
  author={Bardes, Adrien and Garrido, Quentin and Ponce, Jean and Chen, Xinlei and Rabbat, Michael and LeCun, Yann and Assran, Mahmoud and Ballas, Nicolas},
  journal={Transactions on Machine Learning Research},
  year={2024}
}

@inproceedings{chen2020simple,
  title     = {A Simple Framework for Contrastive Learning of Visual Representations},
  author    = {Chen, Ting and Kornblith, Simon and Norouzi, Mohammad and Hinton, Geoffrey},
  booktitle = {Proceedings of the International Conference on Machine Learning (ICML)},
  pages     = {1597--1607},
  year      = {2020}
}

@inproceedings{cho2014gru,
  title     = {Learning Phrase Representations using {RNN} Encoder--Decoder for Statistical Machine Translation},
  author    = {Cho, Kyunghyun and van Merri{\"e}nboer, Bart and Gulcehre, Caglar and Bahdanau, Dzmitry and Bougares, Fethi and Schwenk, Holger and Bengio, Yoshua},
  booktitle = {Proceedings of the Conference on Empirical Methods in Natural Language Processing (EMNLP)},
  year      = {2014}
}

@inproceedings{farneback2003two,
  title     = {Two-Frame Motion Estimation Based on Polynomial Expansion},
  author    = {Farneb{\"a}ck, Gunnar},
  booktitle = {Proceedings of the Scandinavian Conference on Image Analysis (SCIA)},
  year      = {2003}
}

@inproceedings{feichtenhofer2021large,
  title={A Large-Scale Study on Unsupervised Spatiotemporal Representation Learning},
  author={Feichtenhofer, Christoph and Fan, Haoqi and Xiong, Bo and Girshick, Ross and He, Kaiming},
  booktitle={Proceedings of the IEEE/CVF Conference on Computer Vision and Pattern Recognition (CVPR)},
  pages={3298--3308},
  year={2021}
}

@inproceedings{feichtenhofer2022masked,
  title={Masked Autoencoders As Spatiotemporal Learners},
  author={Feichtenhofer, Christoph and Fan, Haoqi and Li, Yanghao and He, Kaiming},
  booktitle={Advances in Neural Information Processing Systems},
  volume={35},
  pages={35946--35958},
  year={2022}
}

@inproceedings{fundel2026moves,
  title     = {What Moves? {L}ocalized Motion Representations for Compositional Scene Control},
  author    = {Fundel, Frank and Ben Alaya, Malek and Ressler-Antal, Thomas and Baumann, Stefan Andreas and Ommer, Bj{\"o}rn},
  booktitle = {Proceedings of the European Conference on Computer Vision (ECCV)},
  pages     = {297--315},
  year      = {2026}
}

@inproceedings{grauman2024egoexo4d,
  title     = {{Ego-Exo4D}: Understanding Skilled Human Activity from First- and Third-Person Perspectives},
  author    = {Grauman, Kristen and Westbury, Andrew and Torresani, Lorenzo and Kitani, Kris and Malik, Jitendra and others},
  booktitle = {Proceedings of the IEEE/CVF Conference on Computer Vision and Pattern Recognition (CVPR)},
  year      = {2024}
}

@inproceedings{he2016deep,
  title     = {Deep Residual Learning for Image Recognition},
  author    = {He, Kaiming and Zhang, Xiangyu and Ren, Shaoqing and Sun, Jian},
  booktitle = {Proceedings of the IEEE/CVF Conference on Computer Vision and Pattern Recognition (CVPR)},
  year      = {2016}
}

@inproceedings{huang2023mgmae,
  title={{MGMAE}: Motion Guided Masking for Video Masked Autoencoding},
  author={Huang, Bingkun and Zhao, Zhiyu and Zhang, Guozhen and Qiao, Yu and Wang, Limin},
  booktitle={Proceedings of the IEEE/CVF International Conference on Computer Vision (ICCV)},
  pages={13447--13458},
  year={2023}
}

@inproceedings{huang2025spacetimepilot,
  title     = {{SpaceTimePilot}: Generative Rendering of Dynamic Scenes Across Space and Time},
  author    = {Huang, Zhening and Jeong, Hyeonho and Chen, Xuelin and Gryaditskaya, Yulia and Wang, Tuanfeng Y. and Lasenby, Joan and Huang, Chun-Hao},
  booktitle = {Proceedings of the IEEE/CVF Conference on Computer Vision and Pattern Recognition (CVPR)},
  pages     = {11218--11228},
  year      = {2026}
}

@inproceedings{kaufmann2023emdb,
  title     = {{EMDB}: The Electromagnetic Database of Global 3D Human Pose and Shape in the Wild},
  author    = {Kaufmann, Manuel and Song, Jie and Guo, Chen and Shen, Kaiyue and Jiang, Tianjian and Tang, Chengcheng and Z{\'a}rate, Juan Jos{\'e} and Hilliges, Otmar},
  booktitle = {Proceedings of the IEEE/CVF International Conference on Computer Vision (ICCV)},
  year      = {2023}
}

@article{knobel2026self,
  title={Self-Supervised Learning of Structured Dynamics from Videos},
  author={Knobel, Lukas and Zisserman, Andrew and Asano, Yuki M},
  journal={arXiv preprint arXiv:2607.21576},
  year={2026}
}

@inproceedings{kubric,
  title     = {Kubric: A Scalable Dataset Generator},
  author    = {Greff, Klaus and Belletti, Francois and Beyer, Lucas and Doersch, Carl and Du, Yilun and Duckworth, Daniel and Fleet, David J. and Gnanapragasam, Dan and Golemo, Florian and Herrmann, Charles and Kipf, Thomas and Kundu, Abhijit and Lagun, Dmitry and Laradji, Issam and Liu, Hsueh-Ti (Derek) and Meyer, Henning and Miao, Yishu and Nowrouzezahrai, Derek and Oztireli, Cengiz and Pot, Etienne and Radwan, Noha and Rebain, Daniel and Sabour, Sara and Sajjadi, Mehdi S. M. and Sela, Matan and Sitzmann, Vincent and Stone, Austin and Sun, Deqing and Vora, Suhani and Wang, Ziyu and Wu, Tianhao and Yi, Kwang Moo and Zhong, Fangcheng and Tagliasacchi, Andrea},
  booktitle = {Proceedings of the IEEE/CVF Conference on Computer Vision and Pattern Recognition (CVPR)},
  pages     = {3749--3761},
  year      = {2022}
}

@inproceedings{leroy2024grounding,
  author    = {Leroy, Vincent and Cabon, Yohann and Revaud, Jerome},
  title     = {Grounding Image Matching in {3D} with {MASt3R}},
  booktitle = {European Conference on Computer Vision},
  pages     = {71--91},
  year      = {2024}
}

@inproceedings{li2018resound,
  title     = {{RESOUND}: Towards Action Recognition without Representation Bias},
  author    = {Li, Yingwei and Li, Yi and Vasconcelos, Nuno},
  booktitle = {Proceedings of the European Conference on Computer Vision (ECCV)},
  year      = {2018}
}

@inproceedings{lin2025camerabench,
  title     = {Towards Understanding Camera Motions in Any Video},
  author    = {Lin, Zhiqiu and Cen, Siyuan and Jiang, Daniel and Karhade, Jay and Wang, Hewei and Mitra, Chancharik and Ling, Yu Tong Tiffany and Huang, Yuhan and Zawar, Rushikesh and Bai, Xue and Du, Yilun and Gan, Chuang and Ramanan, Deva},
  booktitle = {Advances in Neural Information Processing Systems (NeurIPS), Datasets and Benchmarks Track},
  year      = {2025}
}

@inproceedings{ling2024dl3dv,
  title     = {{DL3DV-10K}: A Large-Scale Scene Dataset for Deep Learning-based {3D} Vision},
  author    = {Ling, Lu and Sheng, Yichen and Tu, Zhi and Zhao, Wentian and Xin, Cheng and Wan, Kun and Yu, Lantao and Guo, Qianyu and Yu, Zixun and Lu, Yawen and others},
  booktitle = {Proceedings of the IEEE/CVF Conference on Computer Vision and Pattern Recognition (CVPR)},
  pages     = {22160--22169},
  year      = {2024}
}

@inproceedings{loshchilov2019adamw,
  title     = {Decoupled Weight Decay Regularization},
  author    = {Loshchilov, Ilya and Hutter, Frank},
  booktitle = {International Conference on Learning Representations (ICLR)},
  year      = {2019}
}

@article{oord2018cpc,
  title   = {Representation Learning with Contrastive Predictive Coding},
  author  = {van den Oord, Aaron and Li, Yazhe and Vinyals, Oriol},
  journal = {arXiv preprint arXiv:1807.03748},
  year    = {2018}
}

@inproceedings{qian2021cvrl,
  title={Spatiotemporal Contrastive Video Representation Learning},
  author={Qian, Rui and Meng, Tianjian and Gong, Boqing and Yang, Ming-Hsuan and Wang, Huisheng and Belongie, Serge and Cui, Yin},
  booktitle={Proceedings of the IEEE/CVF Conference on Computer Vision and Pattern Recognition (CVPR)},
  pages={6960--6970},
  year={2021}
}

@inproceedings{recurrentvideo,
  title     = {Recurrent Video Masked Autoencoders},
  author    = {Zoran, Daniel and Parthasarathy, Nikhil and Yang, Yi and Hudson, Drew A. and Carreira, Jo{\~a}o and Zisserman, Andrew},
  booktitle = {Proceedings of the IEEE/CVF Conference on Computer Vision and Pattern Recognition (CVPR)},
  pages     = {17744--17755},
  year      = {2026}
}

@inproceedings{schoenberger2016structure,
  author    = {Sch{\"o}nberger, Johannes L. and Frahm, Jan-Michael},
  title     = {Structure-from-Motion Revisited},
  booktitle = {Proceedings of the IEEE Conference on Computer Vision and Pattern Recognition (CVPR)},
  pages     = {4104--4113},
  year      = {2016}
}

@inproceedings{simonyan2014two,
  title     = {Two-Stream Convolutional Networks for Action Recognition in Videos},
  author    = {Simonyan, Karen and Zisserman, Andrew},
  booktitle = {Advances in Neural Information Processing Systems (NeurIPS)},
  pages     = {568--576},
  year      = {2014}
}

@inproceedings{ssv2,
  title     = {The ``Something Something'' Video Database for Learning and Evaluating Visual Common Sense},
  author    = {Goyal, Raghav and Kahou, Samira Ebrahimi and Michalski, Vincent and Materzynska, Joanna and Westphal, Susanne and Kim, Heuna and Haenel, Valentin and Fruend, Ingo and Yianilos, Peter and Mueller-Freitag, Moritz and others},
  booktitle = {Proceedings of the IEEE International Conference on Computer Vision (ICCV)},
  pages     = {5842--5850},
  year      = {2017}
}

@InProceedings{sturm12iros,
  author = {J. Sturm and N. Engelhard and F. Endres and W. Burgard and D. Cremers},
  title = "A Benchmark for the Evaluation of RGB-D SLAM Systems",
  booktitle = "Proc. of the International Conference on Intelligent Robot Systems (IROS)",
  year = "2012",
  month= "Oct.",
}

@inproceedings{sun2023mme,
  title={Masked Motion Encoding for Self-Supervised Video Representation Learning},
  author={Sun, Xinyu and Chen, Peihao and Chen, Liangwei and Li, Changhao and Li, Thomas H. and Tan, Mingkui and Gan, Chuang},
  booktitle={Proceedings of the IEEE/CVF Conference on Computer Vision and Pattern Recognition (CVPR)},
  pages={2235--2245},
  year={2023}
}

@inproceedings{tan2026natural,
  title     = {Natural Language Camera Movement Understanding},
  author    = {Tan, Yuwen and Huang, Joey and Huang, Jin and Li, Haoxiang and Gong, Boqing},
  booktitle = {European Conference on Computer Vision (ECCV)},
  pages     = {262--280},
  year      = {2026}
}

@inproceedings{teed2021droid,
  author    = {Teed, Zachary and Deng, Jia},
  title     = {{DROID-SLAM}: Deep Visual {SLAM} for Monocular, Stereo, and {RGB-D} Cameras},
  booktitle = {Advances in Neural Information Processing Systems},
  volume    = {34},
  pages     = {16558--16569},
  year      = {2021}
}

@inproceedings{tesch2026bedlam2,
  title     = {{BEDLAM2.0}: Synthetic Humans and Cameras in Motion},
  author    = {Tesch, Joachim and Becherini, Giorgio and Achar, Prerana and Yiannakidis, Anastasios and Kocabas, Muhammed and Patel, Priyanka and Black, Michael J.},
  booktitle = {Advances in Neural Information Processing Systems (NeurIPS), Datasets and Benchmarks Track},
  year      = {2025}
}

@misc{timemachine,
  title         = {The {TIME} Machine: On The Power of Motion for Efficient Perception},
  author        = {Skackauskas, Mantas and Hao, Xinyue and Sevilla-Lara, Laura},
  year          = {2026},
  eprint        = {2605.23045},
  archivePrefix = {arXiv},
  primaryClass  = {cs.CV}
}

@inproceedings{tong2022videomae,
  title={{VideoMAE}: Masked Autoencoders are Data-Efficient Learners for Self-Supervised Video Pre-Training},
  author={Tong, Zhan and Song, Yibing and Wang, Jue and Wang, Limin},
  booktitle={Advances in Neural Information Processing Systems},
  volume={35},
  pages={10078--10093},
  year={2022}
}

@inproceedings{vaswani2017attention,
  title     = {Attention Is All You Need},
  author    = {Vaswani, Ashish and Shazeer, Noam and Parmar, Niki and Uszkoreit, Jakob and Jones, Llion and Gomez, Aidan N. and Kaiser, {\L}ukasz and Polosukhin, Illia},
  booktitle = {Advances in Neural Information Processing Systems (NeurIPS)},
  year      = {2017}
}

@inproceedings{vggtomega,
  title     = {{VGGT-$\Omega$}},
  author    = {Wang, Jianyuan and Chen, Minghao and Zhang, Shangzhan and Karaev, Nikita and Sch{\"o}nberger, Johannes and Labatut, Patrick and Bojanowski, Piotr and Novotny, David and Vedaldi, Andrea and Rupprecht, Christian},
  booktitle = {Proceedings of the IEEE/CVF Conference on Computer Vision and Pattern Recognition (CVPR)},
  pages     = {21486--21499},
  year      = {2026}
}

@inproceedings{videomaev2,
  title     = {{VideoMAE V2}: Scaling Video Masked Autoencoders with Dual Masking},
  author    = {Wang, Limin and Huang, Bingkun and Zhao, Zhiyu and Tong, Zhan and He, Yinan and Wang, Yi and Wang, Yali and Qiao, Yu},
  booktitle = {Proceedings of the IEEE/CVF Conference on Computer Vision and Pattern Recognition (CVPR)},
  pages     = {14549--14560},
  year      = {2023}
}

@article{vjepa2,
  title={V-JEPA 2: Self-Supervised Video Models Enable Understanding, Prediction and Planning},
  author={Assran, Mahmoud and others},
  journal={arXiv preprint arXiv:2506.09985},
  year={2025}
}

@article{vjepa21,
  title   = {{V-JEPA 2.1}: Unlocking Dense Features in Video Self-Supervised Learning},
  author  = {Mur-Labadia, Lorenzo and Muckley, Matthew and Bar, Amir and Assran, Mido and Sinha, Koustuv and Rabbat, Mike and LeCun, Yann and Ballas, Nicolas and Bardes, Adrien},
  journal = {arXiv preprint arXiv:2603.14482},
  year    = {2026}
}

@inproceedings{wang2013action,
  title     = {Action Recognition with Improved Trajectories},
  author    = {Wang, Heng and Schmid, Cordelia},
  booktitle = {Proceedings of the IEEE International Conference on Computer Vision (ICCV)},
  pages     = {3551--3558},
  year      = {2013}
}

@inproceedings{wang2024dust3r,
  author    = {Wang, Shuzhe and Leroy, Vincent and Cabon, Yohann and Chidlovskii, Boris and Revaud, Jerome},
  title     = {{DUSt3R}: Geometric {3D} Vision Made Easy},
  booktitle = {Proceedings of the IEEE/CVF Conference on Computer Vision and Pattern Recognition (CVPR)},
  pages     = {20697--20709},
  year      = {2024}
}

@inproceedings{wang2024motionctrl,
  title     = {{MotionCtrl}: A Unified and Flexible Motion Controller for Video Generation},
  author    = {Wang, Zhouxia and Yuan, Ziyang and Wang, Xintao and Li, Yaowei and Chen, Tianshui and Xia, Menghan and Luo, Ping and Shan, Ying},
  booktitle = {ACM SIGGRAPH Conference Papers},
  year      = {2024}
}

@inproceedings{wang2025continuous,
  author    = {Wang, Qianqian and Zhang, Yifei and Holynski, Aleksander and Efros, Alexei A. and Kanazawa, Angjoo},
  title     = {Continuous {3D} Perception Model with Persistent State},
  booktitle = {Proceedings of the IEEE/CVF Conference on Computer Vision and Pattern Recognition (CVPR)},
  pages     = {10510--10522},
  year      = {2025}
}

@inproceedings{wang2025vggt,
  author    = {Wang, Jianyuan and Chen, Minghao and Karaev, Nikita and Vedaldi, Andrea and Rupprecht, Christian and Novotny, David},
  title     = {{VGGT}: Visual Geometry Grounded Transformer},
  booktitle = {Proceedings of the IEEE/CVF Conference on Computer Vision and Pattern Recognition (CVPR)},
  pages     = {5294--5306},
  year      = {2025}
}

@article{wu2026camreasoner,
  title   = {{CamReasoner}: Reinforcing Camera Movement Understanding via Structured Spatial Reasoning},
  author  = {Wu, Hang and Cai, Yujun and Li, Zehao and Ge, Haonan and Sun, Bowen and Yuan, Junsong and Wang, Yiwei},
  journal = {arXiv preprint arXiv:2602.00181},
  year    = {2026}
}

@inproceedings{xue2026seeing,
  title     = {Seeing without Pixels: Perception from Camera Trajectories},
  author    = {Xue, Zihui and Grauman, Kristen and Damen, Dima and Zisserman, Andrew and Han, Tengda},
  booktitle = {Proceedings of the IEEE/CVF Conference on Computer Vision and Pattern Recognition (CVPR)},
  year      = {2026}
}

@inproceedings{zhang2025monst3r,
  title     = {{MonST3R}: A Simple Approach for Estimating Geometry in the Presence of Motion},
  author    = {Zhang, Junyi and Herrmann, Charles and Hur, Junhwa and Jampani, Varun and Darrell, Trevor and Cole, Forrester and Sun, Deqing and Yang, Ming-Hsuan},
  booktitle = {Proceedings of the International Conference on Learning Representations (ICLR)},
  year      = {2025}
}

@article{zhou2026unveiling,
  title   = {Unveiling the Value of Motion for Cinematic Camera Trajectories},
  author  = {Zhou, Ziqi and Yuan, Yujian and Sevilla-Lara, Laura},
  journal = {arXiv preprint arXiv:2609.38683},
  year    = {2026}
}

@inproceedings{zhu2024languagebind,
  title     = {{LanguageBind}: Extending Video-Language Pretraining to N-modality by Language-based Semantic Alignment},
  author    = {Zhu, Bin and Lin, Bin and Ning, Munan and Yan, Yang and Cui, Jiaxi and Wang, HongFa and Pang, Yatian and Jiang, Wenhao and Zhang, Junwu and Li, Zongwei and others},
  booktitle = {International Conference on Learning Representations (ICLR)},
  year      = {2024}
}
}

\clearpage
\appendix
\section*{Appendix}

\section{\bench{} details}
\label{app:simcambench}

Table~\ref{tab:simcambench-data} summarizes both subsets together with the tasks of Sec.~\ref{sec:bench-eval}, and the
subsections below give the details.
We will release the code, the \bench-Object videos, and the splits, windows and labels of \bench-Human, which can be
applied to the Cam$\times$Time data.
Both subsets also provide training data for the mappings of Sec.~\ref{sec:method}. We split them by room or scene: of
the 500 \bench-Object triplets, 380 in 62 rooms are used for training, 20 in 6 rooms for validation, and 100 in 12 rooms
are held out, together with 101 triplets from a second rendering of the same rooms (Sec.~\ref{app:method-data}). Of the 417 \bench-Human scenes, 300 are used for training, 33 for validation, and 84 are held out. The
held-out rooms and scenes are used neither for training nor for model selection, and the \bench{} results of the
mappings are reported on them.

\begin{table}[!h]
\centering
\footnotesize
\setlength{\tabcolsep}{3.5pt}
\begin{tabular}{@{}lr@{\enspace}r@{\enspace}cl@{\enspace}r@{\enspace}c@{}}
\toprule
 & \multicolumn{3}{c}{Camera motion} & \multicolumn{3}{c}{Scene motion} \\
\cmidrule(lr){2-4}\cmidrule(l){5-7}
Subset & Classes & Clips & Metric & Task & Clips & Metric \\
\midrule
Object & 15 & 500 & mAP & 21 physics families & 500 & acc. \\   %
Human & 11 & 1,667 & mAP & same moment & 5,001 & AUC \\  %
\bottomrule
\end{tabular}
\caption{\bench{} at a glance.}
\label{tab:simcambench-data}
\end{table}

\subsection{\bench-Object}
\label{app:simcambench-object}

\noindent\textbf{Rooms and objects.}
We use 80 rooms based on 5 layouts, varying their colors and adding wall stripes and floor markings. Each clip contains
one to five rigid shapes, counting any static obstacles. We use Kubric's spheres and boxes~\cite{kubric} and custom meshes
for cylinders and capsules. Moving objects have a bounding radius of 0.12--0.51\,m (median 0.26\,m). To vary their
appearance, we use five material profiles (satin, rubber, matte, glossy and brushed metal) and jitter roughness,
specularity and color per object.

\begin{figure*}[t]
\centering
\includegraphics[width=\linewidth]{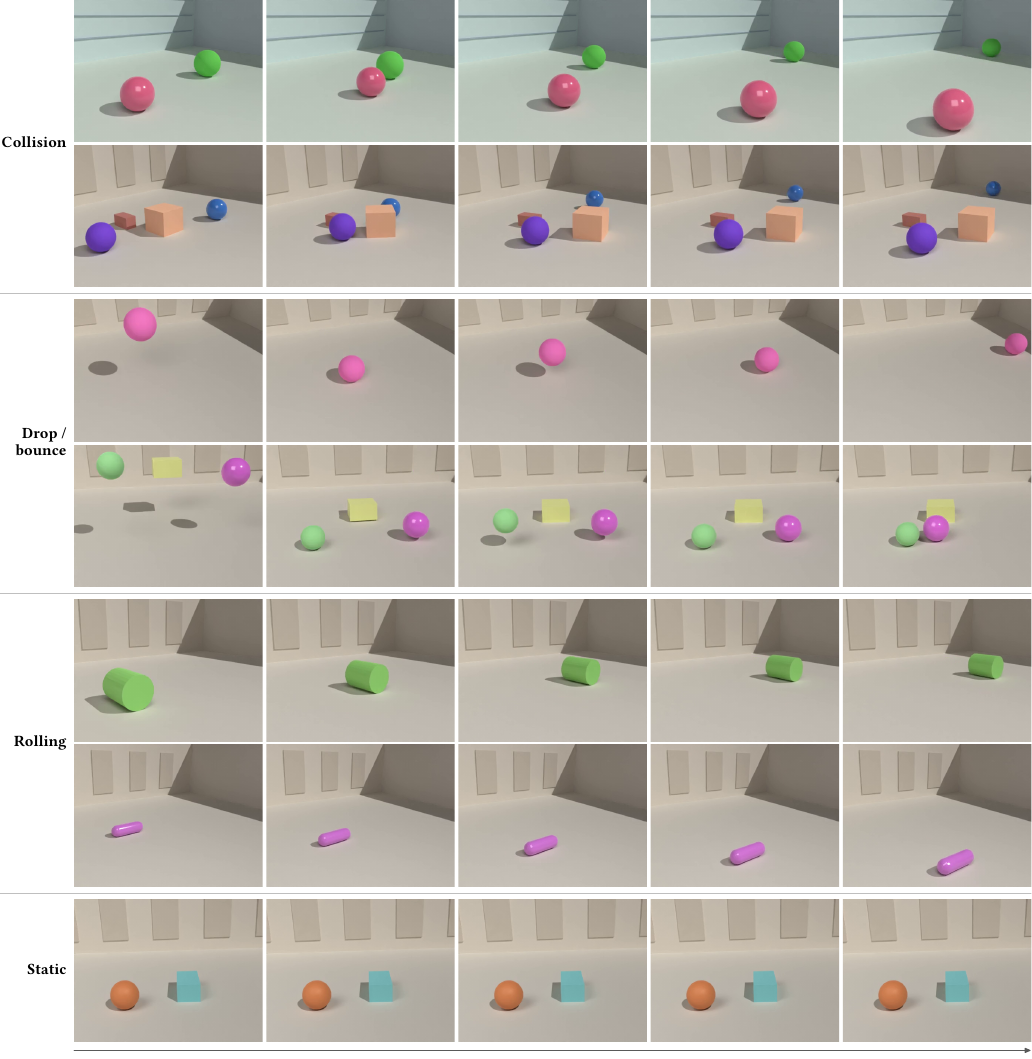}
\caption{Examples of the four physics types of \bench-Object in the camera-frozen version, so that only the
objects move. Each row shows key frames of one clip.}
\label{fig:app-gallery-physics}
\end{figure*}

\vspace{.5em}\noindent\textbf{Physics.}
We simulate each clip with Kubric's PyBullet backend at 240\,Hz under Kubric's default gravity and render it with Blender
(Cycles) at 24\,fps and $512\times384$, which gives clips of 72--120 frames (3--5\,s).
For variety within a family, we sample each object's size, initial position, linear and angular velocity, mass, friction
(0.015--0.82) and restitution (0.12--0.96) from clipped Gaussians.
The four types of physics are collision (332 clips), drop/bounce (84), rolling (55) and static (29), and
Fig.~\ref{fig:app-gallery-physics} shows examples of each.
In collision clips, two to five objects collide in mid-air or near the floor. In drop/bounce clips, objects fall and
bounce, and in rolling clips a cylinder or capsule rolls across the floor. In static clips, a sphere and a box rest
while only the camera moves.
Within these types, we define 21 physics families, each a scenario with its own objects and contacts, such
as two spheres colliding, a sphere falling onto a box, a three-body chain or a sphere hitting a tall static block. The
physics-family task of Sec.~\ref{sec:bench-eval} predicts which of the 21 generated a clip.
Finally, to keep only clips that show the intended physics, every clip must pass contact checks for its family, together
with checks against penetration, floating rebounds and sudden stops.

\vspace{.5em}\noindent\textbf{Camera trajectories.}
All clips draw their start pose from one shared distribution, independent of both the motion and the physics: the camera
looks at a point near the scene center from a horizontal distance of 3.0--5.0\,m, at up to $\pm45^\circ$ azimuth and 0.9--2.6\,m height, with a
focal length of 30--42\,mm.
From there, it moves along seven axes defined in its first-frame coordinates: forward, right, up, tilt, pan, roll and zoom.
Each axis is independently active or near zero, with a random sign, and Table~\ref{tab:app-thresholds} lists the sampled
magnitudes.
In total, 120 clips (24\%) keep the camera still, while the moving clips combine one to seven axes, most often two to four.
Translation follows a straight line and rotation a fixed axis, but the speed along the path varies between linear,
ease-in, ease-out, ease-in-out and two-stage profiles.
We accept a draw only if every moving object stays in view for at least 95\% of the frames with the camera held at its
start pose and if, for both the motion and its mirror image, the camera stays inside the room and every moving object
stays in view for at least 75\% of the frames.
Requiring the mirror image to pass as well keeps the start pose from revealing the direction of motion.
We then check that the start pose reveals little about the motion by probing the camera labels from the start pose alone:
a linear model and a random forest reach 16.6 and 16.1 mAP, against 16.2 and 16.5 when the start poses are shuffled across
clips (mean of 20 shuffles, chance 15.8).

\vspace{.5em}\noindent\textbf{Frozen versions.}
We render each simulation twice more: the camera-frozen version keeps the camera at its first-frame pose and focal length,
and the scene-frozen version replays the camera path with every object held at its first-frame pose.

\begin{figure*}[t]
\centering
\includegraphics[width=\linewidth]{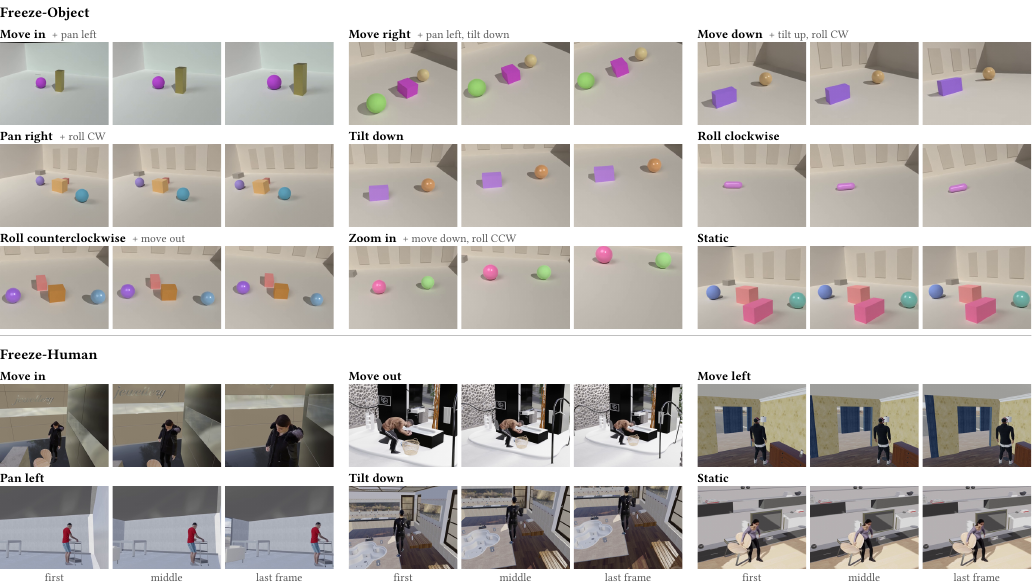}
\caption{Examples of camera primitives in \bench-Object (top) and \bench-Human (bottom): the first, middle and last
frames of the scene-frozen version of a clip, so that only the camera moves. Gray text lists the other camera motions
present in the same clip.}
\label{fig:app-gallery-camera}
\end{figure*}

\vspace{.5em}\noindent\textbf{Camera labels.}
We label each clip with the 15 binary primitives of CameraBench~\cite{lin2025camerabench}: move in/out/left/right/up/down,
pan left/right, tilt up/down, roll clockwise/counter-clockwise, zoom in/out, and static.
Each primitive is computed from the ground-truth change in camera pose and focal length between the first and last frames,
with a positive and a negative threshold (Table~\ref{tab:app-thresholds}).
Because active magnitudes start above the positive threshold and near-zero magnitudes stay below half the negative one, no
clip falls between the two, so every label is positive or negative.
Each moving primitive has 44--94 positive clips, with move up and down the rarest,
and static has 120.
Fig.~\ref{fig:app-gallery-camera} shows examples of the primitives in both subsets of \bench.

\begin{table}[!h]
\centering
\footnotesize
\begin{tabular}{@{}lcccc@{}}
\toprule
 & \multicolumn{2}{c}{Sampled magnitude} & \multicolumn{2}{c}{Label threshold} \\
\cmidrule(lr){2-3}\cmidrule(l){4-5}
Axis & Active & Near zero & Positive & Negative \\
\midrule
Forward (move in/out) & .12--.40 & $\le$.015 & .10 & .03 \\
Right (move right/left) & .12--.45 & $\le$.015 & .10 & .03 \\
Up (move up/down) & .12--.30 & $\le$.015 & .10 & .03 \\
Pan & 6--20$^\circ$ & $\le$0.75$^\circ$ & $5^\circ$ & $1.5^\circ$ \\
Tilt & 6--14$^\circ$ & $\le$0.75$^\circ$ & $5^\circ$ & $1.5^\circ$ \\
Roll & 6--12$^\circ$ & $\le$0.75$^\circ$ & $5^\circ$ & $1.5^\circ$ \\
Zoom & .05--.25 & $\le$.005 & .03 & .01 \\
\bottomrule
\end{tabular}
\caption{Camera motion on \bench-Object between the first and last frames. Active axes are sampled above the positive label
threshold and near-zero axes below half the negative one. Translations are fractions of the camera-to-look-at distance, and
zoom is the change in log focal length.}
\label{tab:app-thresholds}
\end{table}

\subsection{\bench-Human}
\label{app:simcambench-human}
\begin{figure}[t]
\centering
\includegraphics[width=\linewidth]{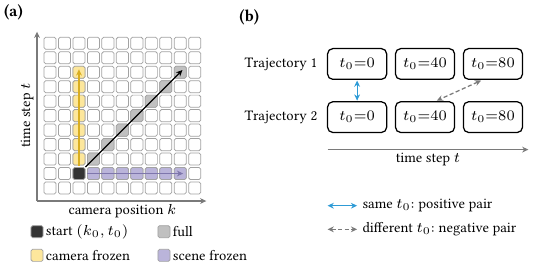}
\caption{\bench-Human from the Cam$\times$Time grid of one trajectory. (a) The full clip and its two frozen versions,
read from the same start $(k_0,t_0)$. (b) The same-moment task: two windows from different trajectories form a
positive pair when they start at the same time step $t_0$. Both windows are taken either in their full or in their
camera-frozen version.}
\label{fig:app-camxtime-grid}
\end{figure}

\begin{figure*}[t]
\centering
\includegraphics[width=\linewidth]{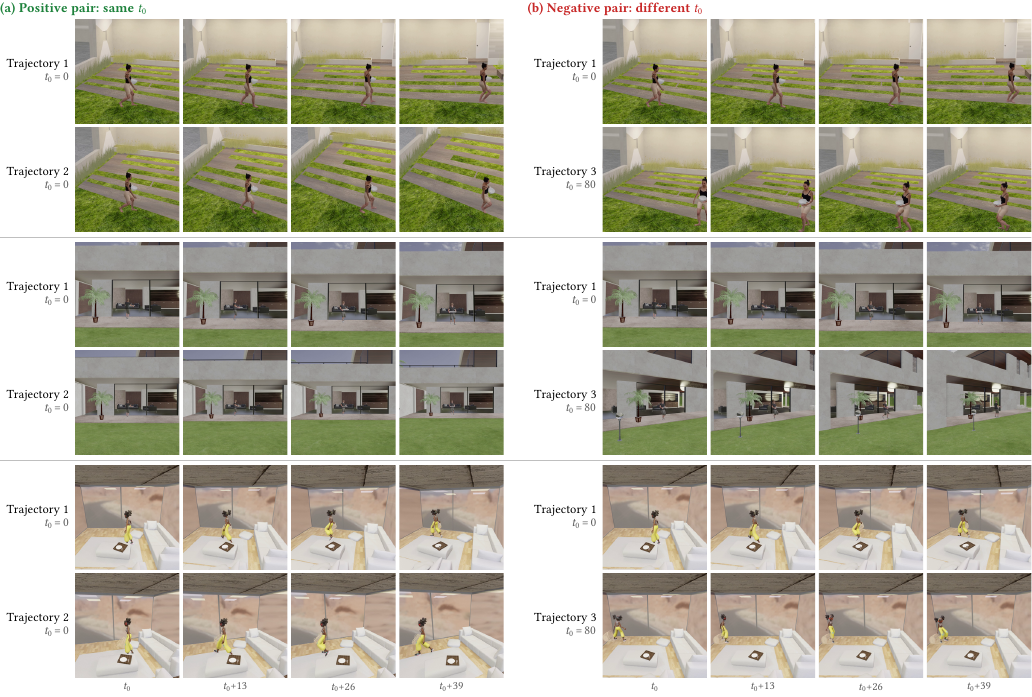}
\caption{Same-moment pairs from three \bench-Human scenes, in the full version. (a) Positive pairs: windows of two
trajectories that start at the same time step $t_0$. (b) Negative pairs: the windows start at different time steps.}
\label{fig:app-gallery-moment}
\end{figure*}

\noindent\textbf{Source data.}
For people, we take the Cam$\times$Time grids~\cite{huang2025spacetimepilot}, which film each animated human character from many
fixed cameras.
We use all 417 scenes of the release available to us, each with four camera trajectories. Along each
trajectory, 120 fixed cameras with known poses film the same 120 time steps of the character's motion at $512^2$ and 30\,fps.
We exclude one trajectory because its camera metadata is incomplete.
Since each trajectory keeps one focal length, zoom never occurs. We also drop the two roll primitives, which the elevation of
the start view predicts (AUC 82\% and 80\%). This leaves 11 of the 15 primitives.

\vspace{.5em}\noindent\textbf{Frozen versions.}
Along each trajectory, the renderings form a grid of camera positions $k$ and time steps $t$, with one image per cell.
A clip is a walk through this grid: from a start $(k_0,t_0)$, the full version moves through cells
$(k_0{+}i,\,t_0{+}i)$, and freezing the camera or the scene fixes one index, giving $(k_0,\,t_0{+}i)$ or
$(k_0{+}i,\,t_0)$ (Fig.~\ref{fig:app-camxtime-grid}a).

\vspace{.5em}\noindent\textbf{Windows.}
Both tasks cut short stretches from these grids, each sampled at 16 frames.
For the camera task, a clip spans 64 grid steps, and we take one per trajectory (1,667 clips) whose first frame shows
the subject near the image center.
For the same-moment task, we instead cut each trajectory into three non-overlapping 40-step windows starting at
$t_0\in\{0,40,80\}$ (5,001 windows).
Each window starts at its own random camera position, which is also the view of its camera-frozen version.
We then pair every two windows of a scene that come from different trajectories and call a pair positive when both start
at the same $t_0$, which gives 22,491 pairs, 7,497 of them positive (Figs.~\ref{fig:app-camxtime-grid}b and~\ref{fig:app-gallery-moment}).

\vspace{.5em}\noindent\textbf{Camera labels.}
We compute the labels of a camera-task clip from the relative pose between its first and last cameras, with the thresholds
of \bench-Object (Table~\ref{tab:app-thresholds}) and translations measured as a fraction of the scene's median
camera-to-subject distance.
Unlike in \bench-Object, these motions are not sampled around the thresholds, so 1,075 of the 18,337 clip--primitive
entries (5.9\%) fall between them, and we leave these out of the average precision.

\subsection{Feature extraction}
\label{app:extra-backbones}
Each backbone receives the 16 frames of a clip and returns 8 temporal steps of dimension $D$ ($D{=}768$, and 384 for
WhatMoves~\cite{fundel2026moves}), whose mean is the clip feature.
For V-JEPA 2.1~\cite{vjepa21} and VideoMAEv2~\cite{videomaev2}, these are the tubelet steps of the last layer, averaged over space.
RVM~\cite{recurrentvideo} gives one mean patch token per frame, which we average over adjacent frame pairs.
WhatMoves gives 13 steps of its global motion stream, which we resample to 8.
Each probe standardizes the clip features and fits a logistic regression ($C{=}1$).
These fits are deterministic, so instead of repeating them with different seeds, we bootstrap their fixed out-of-fold
predictions, resampling triplets for \bench-Object and scenes for \bench-Human 1,000 times.
We report 1.96 bootstrap standard errors of each gap as its 95\% confidence interval.
Table~\ref{tab:app-main-ci} gives these intervals for the gaps of Table~\ref{tab:simcambench-results}.

\begin{table}[t]
\centering
\footnotesize
\setlength{\tabcolsep}{2pt}
\begin{tabular}{@{}l cccc@{}}
\toprule
 & \multicolumn{2}{c}{Camera} & Physics & Same moment \\
 & Object & Human & Object & Human \\
\midrule
V-JEPA 2.1~\cite{vjepa21} & $+5.5{\pm}1.6$ & $+3.1{\pm}1.0$ & $+18.8{\pm}3.7$ & $+14.3{\pm}0.9$ \\
VideoMAEv2~\cite{videomaev2} & $+3.7{\pm}2.5$ & $+8.3{\pm}1.9$ & $+4.8{\pm}2.2$ & $+8.6{\pm}0.9$ \\
RVM~\cite{recurrentvideo} & $+10.1{\pm}2.0$ & $+0.9{\pm}0.5$ & $+12.0{\pm}3.9$ & $+11.3{\pm}1.0$ \\
WhatMoves~\cite{fundel2026moves} & $+12.1{\pm}2.1$ & $+3.2{\pm}1.1$ & $+8.0{\pm}3.2$ & $+18.7{\pm}1.3$ \\
\bottomrule
\end{tabular}
\caption{The gaps $\Delta$ of Table~\ref{tab:simcambench-results} with $\pm$1.96 bootstrap standard errors (points).}
\label{tab:app-main-ci}
\end{table}

\section{Method details}
\label{app:method}

This section gives the configuration of the networks, the computation of the pose targets, the training data and the
training schedule of the two mappings of Sec.~\ref{sec:method}. The video backbones stay fixed throughout.

\subsection{Networks}
\label{app:method-net}

Table~\ref{tab:app-net} lists the layers of the networks of Sec.~\ref{sec:method}.
The networks $w_\psi$ and $u_\phi$ of the camera and scene mappings (Eqs.~\eqref{eq:main-1} and~\eqref{eq:main-5}) share one architecture. Every feature dimension of their input is
standardized with statistics of the training clips, and each Transformer block~\cite{vaswani2017attention} applies
self-attention over all steps.

\vspace{.5em}\noindent\textbf{Pose head.}
Let $H_t \in \mathbb{R}^{512}$ be the output of the Transformer blocks of $h_{\mathrm{pose}}$ at step $t$.
To match the few degrees of freedom of camera motion, $h_{\mathrm{pose}}$ compresses these outputs into a clip code and
a step code for each step,
\begin{equation}
\label{eq:app-codes}
\begin{aligned}
z^{\mathrm{clip}} &= W_{\mathrm{c}}\,\tfrac{1}{T}\textstyle\sum_{t=1}^{T} H_t \in \mathbb{R}^{32}, \\
z^{\mathrm{step}}_t &= W_{\mathrm{s}}\big(H_t + W_{\mathrm{z}}\, z^{\mathrm{clip}}\big) \in \mathbb{R}^{4},
\end{aligned}
\end{equation}
and predicts the directions and log speeds between steps $t$ and $t{+}1$ from the concatenation of the clip code and the
two step codes,
\begin{equation}
\label{eq:app-pose-head}
(\hat d_t, \hat s_t) = W_{\mathrm{o}}\,\big[z^{\mathrm{clip}};\, z^{\mathrm{step}}_t;\, z^{\mathrm{step}}_{t+1}\big],
\end{equation}
for $t = 1, \dots, T{-}1$, where the $W$ are learned linear layers and $[\,\cdot\,;\,\cdot\,]$ denotes concatenation.

\vspace{.5em}\noindent\textbf{Auxiliary networks of the scene mapping.}
The InfoNCE loss~\cite{oord2018cpc} $\mathrm{NCE}$ of Eq.~\eqref{eq:main-8} compares the embeddings of $\varphi$ by cosine similarity with a temperature
of 0.1, and the negatives of a clip are the other clips of its batch.
To preserve scene motion, we train $h_{\mathrm{pred}}$ to predict the next step of the backbone features (Eq.~\eqref{eq:main-9}). The
head runs a GRU~\cite{cho2014gru} over $\text{\scene}_{1:t}$ and predicts $x_{t+1}$ as a correction to $\text{\scene}_t$:
\begin{equation}
\label{eq:app-pred}
\begin{aligned}
\hat x_{t+1} = {} & \text{\scene}_t \\
  & + \mathrm{MLP}\big(\big[W_{\mathrm{p}}\,\mathrm{LN}(\text{\scene}_t);\, o_t\big]\big),
\end{aligned}
\end{equation}
where $W_{\mathrm{p}}$ is a learned linear layer to 256 dimensions, $o_t$ is the state of the GRU after step $t$, projected
to 128 dimensions, and $\mathrm{LN}$ is layer normalization.

\begin{table}[t]
\centering
\small
\setlength{\tabcolsep}{3.5pt}
\begin{tabular}{@{}lll@{}}
\toprule
Layer & Output & Settings \\
\midrule
\multicolumn{3}{@{}l}{\emph{Networks $w_\psi$, $u_\phi$}} \\
Input $x$ (standardized) & $T{\times}D$ & -- \\
Linear + position emb. & $T{\times}512$ & -- \\
Transformer block $\times 4$ & $T{\times}512$ & 8 heads, FF 2048 \\
Linear & $T{\times}D$ & -- \\
\midrule
\multicolumn{3}{@{}l}{\emph{Pose head $h_{\mathrm{pose}}$}} \\
Input \cam{} & $T{\times}D$ & -- \\
Linear + position emb. & $T{\times}512$ & -- \\
Transformer block $\times 4$ & $T{\times}512$ & 8 heads, FF 2048 \\
Clip code: mean, linear & $32$ & -- \\
Step codes: linear & $T{\times}4$ & -- \\
Linear on codes & $(\hat d_t, \hat s_t)_{t<T}$ & -- \\
\midrule
\multicolumn{3}{@{}l}{\emph{Embedding network $\varphi$}} \\
Mean, late minus early half & $2D$ & -- \\
LN & $2D$ & -- \\
MLP & $128$ & 1 hidden layer, 256 \\
\midrule
\multicolumn{3}{@{}l}{\emph{Prediction head $h_{\mathrm{pred}}$}} \\
GRU input: LN + linear & $T{\times}256$ & -- \\
GRU & $T{\times}256$ & 1 layer \\
Linear on GRU state & $T{\times}128$ & -- \\
$W_{\mathrm{p}}$: LN + linear & $T{\times}256$ & -- \\
MLP, Eq.~\eqref{eq:app-pred} & $(T{-}1){\times}D$ & 2 hidden layers, 512 \\
\bottomrule
\end{tabular}
\caption{Layers of the networks. $T{=}8$ steps, $D$ is the feature dimension of the backbone, and LN is layer normalization.}
\label{tab:app-net}
\end{table}

\subsection{Targets of the pose loss}
\label{app:method-cam}

This section describes how we compute the directions $d^m_t$ and log speeds $s^m_t$ of $\mathcal{L}_{\mathrm{pose}}$ in
Eq.~\eqref{eq:main-3} and the steps $\mathcal{S}^m$ over which its direction terms are averaged.
From each training window, we sample 16 frames uniformly and keep the camera pose of every second one, which gives one
pose per step.
Let $R_t$ and $p_t$ be the rotation and position of the camera at step $t$ relative to the first camera, and
$\Delta\tau_t$ the time from step $t$ to $t{+}1$.
The motion between the two steps consists of the displacement $\Delta p_t = p_{t+1} - p_t$ and the rotation vector
$\omega_t$ of the relative rotation $R_t^{\top}R_{t+1}$, which points along the rotation axis and whose length is the
rotation angle in degrees.
Following~\cite{zhou2026unveiling}, we split each into a unit direction and a log speed,
\begin{equation}
\label{eq:app-dirspeed}
\begin{aligned}
d^{\mathrm{tr}}_t &= \frac{\Delta p_t}{\|\Delta p_t\|}, & s^{\mathrm{tr}}_t &= \log\frac{\|\Delta p_t\|}{\Delta\tau_t}, \\
d^{\mathrm{rot}}_t &= \frac{\omega_t}{\|\omega_t\|}, & s^{\mathrm{rot}}_t &= \log\frac{\|\omega_t\|}{\Delta\tau_t}.
\end{aligned}
\end{equation}
We add $10^{-6}$ to each speed before the logarithm, so that both log speeds stay finite for a still camera.
Poses estimated from monocular video are known only up to a scale per clip, which shifts all $s^{\mathrm{tr}}_t$ of a
clip by the same constant. We therefore subtract, for every clip, the mean of $s^{\mathrm{tr}}_t$ over its steps. Both log speeds are then
standardized with training statistics.
A step belongs to $\mathcal{S}^{\mathrm{tr}}$ if its translation is at least the 10th percentile of the step lengths in the
training windows, and to $\mathcal{S}^{\mathrm{rot}}$ if it turns by at least $0.1^\circ$. A window without such steps
contributes only to the speed terms.
 The Huber loss $\ell_{\mathrm{H}}$ uses the threshold $\delta{=}1$.

\subsection{Training data}
\label{app:method-data}

\noindent\textbf{Synthetic pairs.}
From each of the 1,199 trajectories of the 300 Cam$\times$Time training scenes (Sec.~\ref{app:simcambench}), we cut ten
windows of 64 or 40 grid steps, which gives 11,990 clips for paired supervision. Each window and its camera-frozen and
scene-frozen versions share the same start. For \bench-Object, we expand the 380 benchmark training triplets to 800 by
adding 420 triplets from a second rendering of simulations not used in the benchmark. This rendering uses the same
pipeline and provides 73 further triplets for validation and the 101 held-out triplets of Sec.~\ref{app:simcambench}. We keep the held-out rooms and scenes out of both training and
validation.

\vspace{.5em}\noindent\textbf{Camera motion.}
The real clips are 52,775 shots from 6,346 videos of Condensed Movies~\cite{bain2020condensed}, split by video into
47,636 clips for training and 5,139 for validation. Their camera poses are estimated by
\mbox{VGGT-$\Omega$}~\cite{vggtomega}.
BEDLAM2~\cite{tesch2026bedlam2} adds 27,454 rendered sequences with ground-truth camera poses, 24,709 for training and
2,745 for validation.

\vspace{.5em}\noindent\textbf{Crop pairs.}
We select clips with a nearly still camera from the training split of Something-Something v2~\cite{ssv2}, excluding
every video used in our evaluations. A clip counts as nearly still if, over 8 of its frames, the mean rotation per step in
degrees plus the mean step of the camera center that VGGT-$\Omega$ estimates stays below 0.5.
We keep at most 100 clips per action template, which leaves 5,332 clips, 4,799 for training and 533 for validation.
Each of these clips replays the per-step pan, tilt, roll and zoom that VGGT-$\Omega$ estimates on a moving clip of the same
split. The moving crop starts as a centered crop covering 75\% of the frame and shrinks where needed to stay inside the frame. The fixed
crop $\bar x$ is the same centered crop without motion.

\vspace{.5em}\noindent\textbf{Next-step prediction.}
$h_{\mathrm{pred}}$ is trained on 5,700 further training clips of Something-Something v2, drawn regardless of their
camera motion and without their labels, and 300 more clips serve for validation.

\subsection{Training}
\label{app:method-train}

We train both mappings with AdamW~\cite{loshchilov2019adamw} (learning rate $3{\times}10^{-4}$, weight decay 0.05),
without a learning-rate schedule.
Each step of the camera mapping uses a batch of 64 synthetic pairs and a batch of 64 movie and 64 BEDLAM2 windows.
Each step of the scene mapping uses a batch of 64 synthetic pairs, one of 64 crop pairs and one of 64 clips for
next-step prediction.
We select the camera mapping using the pose loss on validation windows from Condensed Movies and BEDLAM2. For the scene
mapping, we use the validation loss on synthetic and crop pairs. In both cases, we validate every 400 steps, stop when the
best validation loss has improved by less than 1\% over the last 10 validations, at the latest after 36,000 and 16,000
steps, and keep the checkpoint with the lowest validation loss.
Following Sec.~\ref{sec:method}, we divide the initial validation loss of the other terms by the initial $\mathcal{L}_{\mathrm{inv}}$ and
$\mathcal{L}_{\mathrm{pred}}$. If either ratio lies outside $[0.1, 10]$, each weight is set to its ratio, and otherwise both
are 1. Computed once per seed before training, this gives $\lambda_{\mathrm{inv}} = \lambda_{\mathrm{pred}} = 1$ for
V-JEPA 2.1, WhatMoves and VideoMAEv2, and $\lambda_{\mathrm{inv}} = 0.029$--$0.035$ and $\lambda_{\mathrm{pred}} =
0.31$--$0.34$ over the three seeds for RVM.
Each run uses one NVIDIA A100 GPU. A run of the camera mapping takes on average 5.3\,h for V-JEPA 2.1, 0.4\,h for
WhatMoves, 2.2\,h for VideoMAEv2 and 2.1\,h for RVM. A run of the scene mapping takes less than an hour.

\section{Experiment details}
\label{app:exp}

This section describes the protocol of every task in Sec.~\ref{sec:experiments} and adds further results.

\subsection{Evaluation protocol}
\label{app:exp-protocol}

\noindent\textbf{Inputs and probes.}
We use the backbone features described in Sec.~\ref{app:extra-backbones} and the probes listed in
Table~\ref{tab:app-exp-tasks}. To compare the original and mapped representations under the same conditions, we use the
same training data and probe hyperparameters for each task.

\vspace{.5em}\noindent\textbf{Seeds.}
Each mapping is trained with three seeds, and each seed gets its own probe. We report the mean over seeds of the per-seed
difference to the backbone features.

\begin{table*}[t]
\centering
\scriptsize
\setlength{\tabcolsep}{3pt}
\begin{tabular}{@{}l l l l@{}}
\toprule
Task & Training $\rightarrow$ test data & Probe (readout) & Metric \\
\midrule
\multicolumn{4}{@{}l}{\emph{Camera understanding}} \\
CameraBench~\cite{lin2025camerabench} & 1,261 labeled training videos $\rightarrow$ 7,623 test records of 963 videos & LR per primitive & mAP over 15 primitives \\
Skill-based retrieval~\cite{lin2025camerabench} & 1,387 training videos $\rightarrow$ 2,431 skill items of 1,065 videos & linear head to text space & group score \\
ACaM~\cite{tan2026natural} & 4,214 training videos $\rightarrow$ 1,375 test items of 1,334 videos & 17-way LR & mean 4-option acc.\ over 17 \\
\bench-O. & 201 triplets of 12 held-out rooms, five folds by room & LR per primitive & mAP over 15 primitives \\
\bench-H. & 336 windows of 84 held-out scenes, five folds by scene & LR per primitive & mAP over 11 primitives \\
\midrule
\multicolumn{4}{@{}l}{\emph{Geometric grounding}} \\
DL3DV~\cite{ling2024dl3dv} & 14,681 $\rightarrow$ 6,295 windows, split by scene (2,099 test scenes) & linear (SGD) & zMSE \\
EMDB~\cite{kaufmann2023emdb} & 1,636 windows of 81 sequences, five folds by sequence & ridge & zMSE \\
TUM RGB-D~\cite{sturm12iros} & 148 windows of 9 dynamic sequences, leave one sequence out & ridge & zMSE \\
\midrule
\multicolumn{4}{@{}l}{\emph{Scene understanding}} \\
\bench-O. & 201 triplets of 12 held-out rooms, five folds by room & 21-way LR (mean) & accuracy \\
\bench-H. & 4,536 window pairs of 84 held-out scenes, five folds by scene & LR on $|x_a - x_b|$ (mean) & ROC-AUC \\
Diving48~\cite{li2018resound} & 15,027 $\rightarrow$ 1,970 clips (official split, 47 classes in the test split) & multinomial LR (concat.) & accuracy \\
SSv2~\cite{ssv2} & 50,516 class-balanced training clips $\rightarrow$ 24,673 validation clips & multinomial LR (concat.) & accuracy \\
EPIC-K.~\cite{Damen2022RESCALING} & 10,325 low-motion training clips $\rightarrow$ all 7,388 validation clips & multinomial LR (concat.) & class-bal.\ accuracy \\
Ego-Exo4D~\cite{grauman2024egoexo4d} & 3,995 low-motion training segments $\rightarrow$ all 3,666 validation segments & multinomial LR (concat.) & class-bal.\ accuracy \\
\bottomrule
\end{tabular}
\caption{Protocol of each task. LR: logistic regression on standardized features ($C{=}1$). \bench-O./H.: \bench-Object/Human.
EPIC-K.: EPIC-KITCHENS. The scene readouts use the mean over the 8 steps (mean) or their concatenation (concat.).}
\label{tab:app-exp-tasks}
\end{table*}

\subsection{Camera tasks}
\label{app:exp-camera}

\vspace{.5em}\noindent\textbf{CameraBench.}
We use the training labels released by the CameraBench authors~\cite{lin2025camerabench} for the 15 camera-motion
primitives, which cover 1,261 training videos.
For each primitive, we train a logistic regression.
We report the mean average precision over the primitives on the official binary test subsets.
We score the 7,623 of the 7,679 test records whose videos are available for every backbone.

\vspace{.5em}\noindent\textbf{Skill-based retrieval.}
We train a linear head to map video representations into the frozen text space of
LanguageBind~\cite{zhu2024languagebind}, using the yes/no questions and captions from the CameraBench training set. We then
report the group score using the authors' evaluator.

\vspace{.5em}\noindent\textbf{ACaM.}
Following ACaM~\cite{tan2026natural}, the probe is trained on its training set, from which we draw at most 300 videos
per class (4,214 videos), and tested on its real-world test set.
At test time, the 17-way probe chooses among the four offered movements, and we report the mean of the 17 per-movement
accuracies, computed with the authors' code.
We use 1,334 of the 1,423 real test videos. The 79 CineTechBench items are unavailable, and we drop 10 that appear in the
CameraBench training set.

\vspace{.5em}\noindent\textbf{Camera pose.}
We evaluate geometric grounding by predicting the relative 6-DoF camera pose between consecutive steps. On
DL3DV~\cite{ling2024dl3dv}, we follow the linear protocol of Knobel~\etal~\cite{knobel2026self} and predict the pose of the
final step pair. EMDB~\cite{kaufmann2023emdb} and TUM RGB-D~\cite{sturm12iros} contain far fewer sequences, so we use ridge
regression on these datasets to predict the pose of every step pair. For all three datasets, we report zMSE, the mean
squared error after target standardization, so that predicting the mean gives 1.

\subsection{Scene tasks}
\label{app:exp-scene}

\noindent\textbf{Held-out \bench{} scenes.}
Both tasks follow Sec.~\ref{sec:simcambench} within the scenes held out from every training set of the mappings: the
physics family on 201 triplets of 12 rooms, of which 101 come from a second rendering of the same rooms, and same moment
on the 4,536 pairs of windows from different trajectories in 84 scenes.

\vspace{.5em}\noindent\textbf{Diving48 and SSv2.}
On Diving48~\cite{li2018resound}, we use the official split of its second version. On SSv2~\cite{ssv2}, the probe is
trained on 50,516 class-balanced training clips and tested on all validation clips available for every backbone.

\vspace{.5em}\noindent\textbf{Egocentric videos.}
On EPIC-KITCHENS-100~\cite{Damen2022RESCALING}, we use the narrated segments of at least 1\,s and the 18 verbs with at least
200 such segments in the videos of the earlier EPIC-KITCHENS-55 release.
We estimate head motion from optical-flow magnitude~\cite{farneback2003two}.
On Ego-Exo4D~\cite{grauman2024egoexo4d}, we use the egocentric view, the keystep segments of at least 1\,s and the 170
keysteps with at least 20 training segments, a lower threshold for this smaller dataset. Its head motion is the mean
angular speed of the camera from the recorded camera poses.
On both datasets, we split the training clips of each class into thirds by head motion and train the probe on the lowest
third. We apply the same thresholds to the validation clips and evaluate on all validation clips (All) and the resulting
high-motion group (High).
The lowest training third is unbalanced, with 90 to 1,954 clips per verb on EPIC-KITCHENS. On both datasets, we therefore
report class-balanced accuracy, so that the most frequent classes do not dominate.

\begin{figure}[t]
\centering
\includegraphics[width=\linewidth]{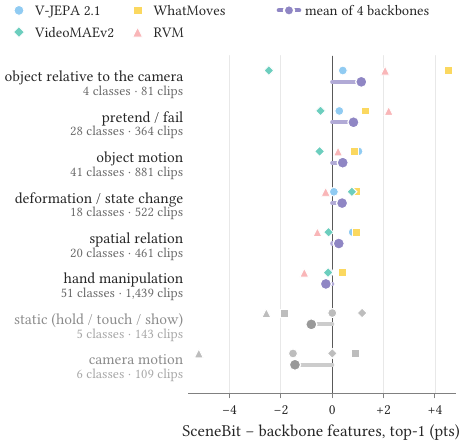}
\caption{Change in SSv2 top-1 accuracy from the backbone features to \scene{}, per category of class labels. We train the
linear probes on 6,005 class-stratified training clips and evaluate on 4,000 randomly sampled validation clips.}
\label{fig:app-ssv2-main}
\end{figure}

\begin{figure*}[t]
\centering
\includegraphics[width=\textwidth]{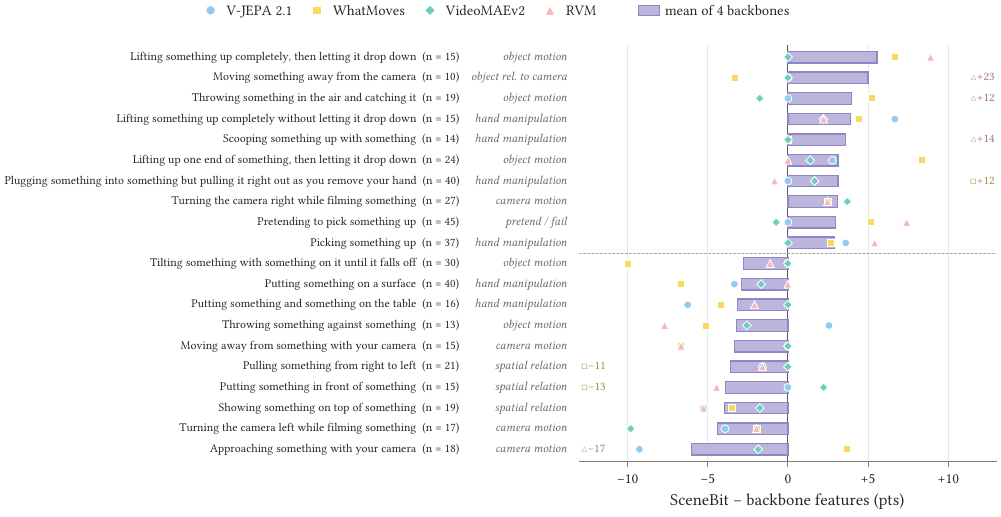}
\caption{The ten SSv2 classes that \scene{} improves most and the ten it hurts most (purple: mean of the four backbones,
markers: individual backbones). The category of each class is given in italics.}
\label{fig:app-ssv2-classes}
\end{figure*}

\subsection{Further results}
\label{app:exp-extra}

\noindent\textbf{Reading camera motion from \scene{}.}
We use the CameraBench linear probes of Sec.~\ref{app:exp-camera} to test how the scene mapping affects the readout of
camera motion. \scene{} scores 8.2, 14.0, 6.3 and 14.7 points of mAP below the backbone features for V-JEPA 2.1,
WhatMoves, VideoMAEv2 and RVM. Together with the scene gains in Table~\ref{tab:scene-real}, these results suggest
that \scene{} can improve scene understanding while making camera motion harder to read, consistent with the interference
observed in Sec.~\ref{sec:simcambench}.

\vspace{.5em}\noindent\textbf{Gap to the camera-frozen clips.}
Since we train the scene mapping to approximate camera-frozen features, we next measure how much of their advantage
\scene{} recovers on the physics-family task. On the held-out \bench-Object rooms, \scene{} closes 86\%, 80\%, 71\% and 75\% of the gap between the backbone features
of the full clips and those of their camera-frozen versions in physics-family accuracy, for V-JEPA 2.1, WhatMoves,
VideoMAEv2 and RVM.

\vspace{.5em}\noindent\textbf{Attentive probe.}
We also evaluate \scene{} with a stronger, attentive probe on EPIC-KITCHENS, where the camera moves with the head.
\scene{} gains 2.2 and 3.1 points on All and High with V-JEPA 2.1, and 1.3 and 2.6 points with WhatMoves
(Table~\ref{tab:app-attentive}), more than with the linear probe.
The attentive probe attends over the 8 steps with one learned query, followed by an MLP and a linear classifier, and is
trained for 20 epochs with the learning rate chosen on 10\% of the training clips.

\begin{table}[t]
\centering
\footnotesize
\setlength{\tabcolsep}{2.5pt}
\begin{tabular}{@{}l cccc@{}}
\toprule
 & V-JEPA 2.1~\cite{vjepa21} & WhatMoves~\cite{fundel2026moves} & VideoMAEv2~\cite{videomaev2} & RVM~\cite{recurrentvideo} \\
\midrule
All & $+2.2$ & $+1.3$ & $+0.1$ & $+0.0$ \\
High & $+3.1$ & $+2.6$ & $-0.1$ & $+0.6$ \\
\bottomrule
\end{tabular}
\caption{Attentive probe: change from the backbone features to \scene{} on all clips (All) and the
high-motion subset (High, Sec.~\ref{app:exp-scene}) of EPIC-KITCHENS~\cite{Damen2022RESCALING} (class-balanced accuracy in points, mean of
three seeds).}
\label{tab:app-attentive}
\end{table}

\subsection{Per-category analysis on SSv2}
\label{app:exp-analysis}

To understand which actions benefit from \scene{}, we group the SSv2 classes by what their labels describe and compare
their accuracy before and after the scene mapping (Fig.~\ref{fig:app-ssv2-main}). Averaged over the four backbones,
classes in which an object moves relative to the camera gain the most. Categories involving object motion, deformation
and pretended actions also gain. By contrast, classes defined by camera motion drop the most, followed by static classes such as holding or touching an object.
In these clips the object barely moves, so a larger share of the motion they contain likely comes from the camera. This pattern is consistent
with Sec.~\ref{sec:exp-analysis}: \scene{} makes camera motion less accessible, which favors actions defined by object motion over those
defined by camera motion.

Fig.~\ref{fig:app-ssv2-classes} shows the classes that change most.

\section{Ablations}
\label{app:ablations}

To examine which components contribute to the gains, we ablate the training terms and three design choices of Sec.~\ref{sec:method}:
the residual form, the Transformer architecture, and separate networks for the two mappings. Each variant changes one
component and otherwise follows the training and evaluation protocols of Secs.~\ref{app:method} and~\ref{app:exp}. We run
these comparisons on WhatMoves with three seeds. Tables~\ref{tab:app-camera-ablations} and~\ref{tab:app-scene-ablations}
report changes relative to the full camera and scene mappings, respectively.

\newcommand{\lsc}[1]{\textcolor{losscolor}{#1}}
\begin{table}[t]
\centering
\footnotesize
\setlength{\tabcolsep}{0.9pt}
\begin{tabular}{@{}l cccc@{}}
\toprule
 & CameraBench~\cite{lin2025camerabench} & \bench-O. & \bench-H. & TUM~\cite{sturm12iros} \\
 & mAP & mAP & mAP & zMSE \\
\midrule
No residual, Eq.~\eqref{eq:main-1} & \lsc{$-0.5$} & \lsc{$-2.5$} & \lsc{$-0.7$} & \lsc{$+.050$} \\
No pair term, Eq.~\eqref{eq:main-2} & \lsc{$-0.5$} & \lsc{$-4.6$} & \lsc{$-0.1$} & $-.008$ \\
No fixed point, Eq.~\eqref{eq:main-2} & $+0.3$ & \lsc{$-0.2$} & \lsc{$-0.7$} & \lsc{$+.021$} \\
No $\mathcal{L}_{\mathrm{pose}}$, Eq.~\eqref{eq:main-3} & \lsc{$-24.3$} & $+1.1$ & \lsc{$-2.2$} & \lsc{$+.585$} \\
Shared network & \lsc{$-1.4$} & $+0.3$ & \lsc{$-0.9$} & \lsc{$+.110$} \\
MLP & \lsc{$-0.7$} & \lsc{$-0.9$} & \lsc{$-0.7$} & \lsc{$+.309$} \\
\bottomrule
\end{tabular}
\caption{Ablations of the camera mapping on WhatMoves~\cite{fundel2026moves}: change relative to the full camera mapping (mean of three seeds).
Red marks a worse score than the full mapping. TUM: TUM RGB-D, for whose zMSE lower is better. Shared network: one network for both mappings. MLP: the Transformer replaced by an MLP with
as many parameters.}
\label{tab:app-camera-ablations}
\end{table}

\begin{table}[t]
\centering
\small
\setlength{\tabcolsep}{3.5pt}
\begin{tabular}{@{}l cc@{}}
\toprule
 & Diving48~\cite{li2018resound} & EPIC-K.~\cite{Damen2022RESCALING} \\
 & Acc. & Acc.\ (All) \\
\midrule
No residual, Eq.~\eqref{eq:main-5} & \lsc{$-2.1$} & \lsc{$-1.7$} \\
No $\mathcal{L}^{\mathrm{scene}}_{\mathrm{syn}}$, Eq.~\eqref{eq:main-6} & \lsc{$-1.1$} & \lsc{$-0.4$} \\
No crop pair term, Eq.~\eqref{eq:main-7} & \lsc{$-0.2$} & $\phantom{+}0.0$ \\
No crop fixed point, Eq.~\eqref{eq:main-7} & \lsc{$-0.7$} & \lsc{$-0.3$} \\
No $\mathcal{L}_{\mathrm{inv}}$, Eq.~\eqref{eq:main-8} & \lsc{$-1.0$} & $\phantom{+}0.0$ \\
No $\mathcal{L}_{\mathrm{pred}}$, Eq.~\eqref{eq:main-9} & $+0.3$ & \lsc{$-0.5$} \\
Shared network & \lsc{$-0.3$} & \lsc{$-1.1$} \\
MLP & \lsc{$-0.6$} & \lsc{$-0.5$} \\
\bottomrule
\end{tabular}
\caption{Ablations of the scene mapping on WhatMoves~\cite{fundel2026moves}: change relative to \scene{} (mean of three seeds, points). Red marks a
worse score. Shared network: one network for both mappings. MLP: the Transformer replaced by an MLP with as many parameters.}
\label{tab:app-scene-ablations}
\end{table}

\vspace{.5em}\noindent\textbf{Camera mapping.}
Without $\mathcal{L}_{\mathrm{pose}}$, CameraBench mAP drops by 24.3 points and the TUM RGB-D error grows by .585. Pose
supervision thus appears to be the main source of the gains on real videos. The pair term of Eq.~\eqref{eq:main-2} also improves
\bench-Object mAP by 4.6 points. Beyond these training signals, the residual form of Eq.~\eqref{eq:main-1} improves performance on
\bench-Object and TUM RGB-D, while using separate networks improves CameraBench and TUM RGB-D. Finally, replacing the
Transformer network with an MLP with the same number of parameters increases the TUM RGB-D error by .309, supporting our
choice of the Transformer architecture.

\vspace{.5em}\noindent\textbf{Scene mapping.}
Removing the residual form of Eq.~\eqref{eq:main-5} produces the largest accuracy drops in Table~\ref{tab:app-scene-ablations}: 2.1
points on Diving48 and 1.7 points on EPIC-KITCHENS. The training terms also contribute: removing any one of them lowers
accuracy on at least one of the two datasets. Finally, sharing a single network or replacing the Transformer network with an
MLP lowers accuracy on both datasets.

Overall, removing or replacing components of either mapping worsens most of the reported results, supporting the design
choices of Sec.~\ref{sec:method}.

\section{Discussion}
\label{app:future-directions}

Camera motion is common in real videos, from handheld and egocentric recordings to films. Our results show that camera
and scene motion interfere in the representations of current video foundation models, and that lightweight mappings on
fixed features can reduce this interference without re-training the backbones. We chose this post-hoc setting because
it applies directly to existing models at a small cost. These results motivate us to investigate whether incorporating
the triplets of \bench{} into backbone training can help the models distinguish camera motion from scene motion more
effectively. Since our pipeline generates the frozen versions automatically, \bench{} can be scaled for this purpose as
resources allow.

Our analysis also shows that the right treatment of camera motion depends on the video and the task, since some actions
are defined by the camera motion itself (Sec.~\ref{sec:exp-analysis}). Representations that adapt to the video and the task are therefore a
promising direction. Beyond the tasks studied here, separating the two motions may also benefit other settings where
camera motion is pervasive, such as depth estimation, tracking and learning from egocentric video.

More broadly, we hope that our benchmark and mappings open a new line of work towards video representations that
understand both how the world moves and how it is filmed.

\section{Ethical considerations}
\label{app:ethics}

This work involves no human subjects research: we collect no new data from people. \bench-Object is fully synthetic,
and \bench-Human and BEDLAM2 use rendered characters. The real videos come from widely used public research datasets,
cited in Secs.~\ref{app:method} and~\ref{app:exp}, which we use under their terms of use and none of which has been
withdrawn by its creators. Our mappings act on the features of fixed video backbones and do not target the identity or other
personal attributes of the people in the videos.

\end{document}